# MUSE CSP: An Extension to
# the Constraint Satisfaction Problem

**Randall A. Helzerman**                              HELZ@ECN.PURDUE.EDU
**Mary P. Harper**                                  HARPER@ECN.PURDUE.EDU
*School of Electrical and Computer Engineering*
*1285 Electrical Engineering Building*
*Purdue University*
*West Lafayette, IN 47907-1285 USA*

## Abstract

This paper describes an extension to the constraint satisfaction problem (CSP) called MUSE CSP (*MU*ltiply *SE*gmented *C*onstraint *S*atisfaction *P*roblem). This extension is especially useful for those problems which segment into multiple sets of partially shared variables. Such problems arise naturally in signal processing applications including computer vision, speech processing, and handwriting recognition. For these applications, it is often difficult to segment the data in only one way given the low-level information utilized by the segmentation algorithms. MUSE CSP can be used to compactly represent several similar instances of the constraint satisfaction problem. If multiple instances of a CSP have some common variables which have the same domains and constraints, then they can be combined into a single instance of a MUSE CSP, reducing the work required to apply the constraints. We introduce the concepts of MUSE node consistency, MUSE arc consistency, and MUSE path consistency. We then demonstrate how MUSE CSP can be used to compactly represent lexically ambiguous sentences and the multiple sentence hypotheses that are often generated by speech recognition algorithms so that grammar constraints can be used to provide parses for all syntactically correct sentences. Algorithms for MUSE arc and path consistency are provided. Finally, we discuss how to create a MUSE CSP from a set of CSPs which are labeled to indicate when the same variable is shared by more than a single CSP.

## 1. Introduction

This paper describes an extension to the constraint satisfaction problem (CSP) called MUSE CSP (*MU*ltiply *SE*gmented *C*onstraint *S*atisfaction *P*roblem). This extension is especially useful for those problems which segment into multiple sets of partially shared variables. First, we describe the constraint satisfaction problem and then define our extension.

### 1.1 The Constraint Satisfaction Problem

Constraint satisfaction problems (CSP) have a rich history in Artificial Intelligence (Davis & Rosenfeld, 1981; Dechter, Meiri, & Pearl, 1991; Dechter & Pearl, 1988; Freuder, 1989, 1990; Mackworth, 1977; Mackworth & Freuder, 1985; Villain & Kautz, 1986; Waltz, 1975) (for a general reference, see Tsang, 1993). Constraint satisfaction provides a convenient way to represent and solve certain types of problems. In general, these are problems which can be solved by assigning mutually compatible values to a predetermined number of variables under a set of constraints. This approach has been used in a variety of disciplines including machine vision, belief maintenance, temporal reasoning, graph theory, circuit design, and diagnostic reasoning. When using a CSP approach (e.g., Figure 1), the variables are typically depicted as vertices or nodes, where each node is associated with a finite set of possible values, and the constraints imposed on the variables are depicted using arcs. An arc looping from a node to itself represents a unary constraint (a constraint on a single variable), and an arc between two nodes represents a binary constraint (a constraint on two variables). A classic example of a CSP is the map coloring problem (e.g., Figure 1), where a color must be assigned to each country such that no two neighboring countries have the same color. A variable represents a country's color, and a constraint arc between two variables indicates that the two joined countries are adjacent and should not be assigned the same color.

Formally, a CSP (Mackworth, 1977) is defined in Definition 1.

**Definition 1** (Constraint Satisfaction Problem)
$N = \{i, j, \ldots\}$ *is the set of* nodes *(or variables), with* $|N| = n$,
$L = \{a, b, \ldots\}$ *is the set of* labels, *with* $|L| = l$,
$L_i = \{a | a \in L \text{ and } (i, a) \text{ is admissible}\}$,
*R1 is a* unary constraint, *and* $(i, a)$ *is admissible if* $R1(i, a)$,
*R2 is a* binary constraint, $(i, a) - (j, b)$ *is admissible if* $R2(i, a, j, b)$.

A CSP network contains all $n$-tuples in $L^n$ which satisfy $R1$ and $R2$. Since some of the labels associated with a node may be incompatible with labels assigned to other nodes, it is desirable, when the constraints are sufficiently tight (van Beek, 1994), to eliminate as many of these labels as possible by enforcing local consistency conditions before a globally consistent solution is extracted (Dechter, 1992). Node and arc consistency are defined in Definitions 2 and 3, respectively. In addition, it may be desirable to eliminate as many label pairs as possible using path consistency, which is defined in Definition 4.

**Definition 2** (Node Consistency) *An instance of CSP is said to be* node consistent *if and only if each node's domain contains only labels for which the unary constraint R1 holds, i.e.:*

$$\forall i \in N : \forall a \in L_i : R1(i, a)$$

**Definition 3** (Arc Consistency) *An instance of CSP is said to be* arc consistent *if and only if for every pair of nodes i and j, each element of $L_i$ (the domain of i) has at least one element of $L_j$ for which the binary constraint R2 holds, i.e.:*

$$\forall i, j \in N : \forall a \in L_i : \exists b \in L_j : R2(i, a, j, b)$$

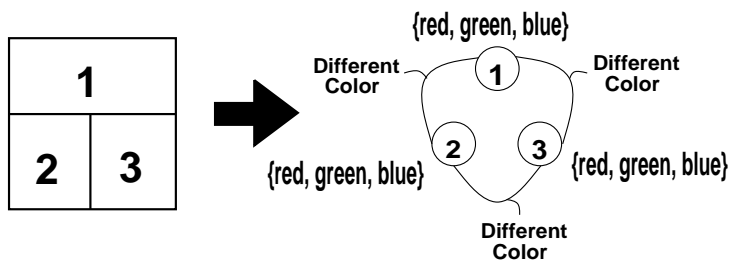

Figure 1: The map coloring problem as an example of CSP.

**Definition 4** (Path Consistency) *An instance of CSP is said to be* path consistent *if and only if:*

$$\forall i,j \in N : i \neq j \Rightarrow (\forall a \in L_i : \forall b \in L_j : \forall k \in N : k \neq i \wedge k \neq j \wedge Path(i,k,j) \Rightarrow$$
$$(R2(i,a,j,b) \Rightarrow \exists c \in L_k : R2(i,a,k,c) \wedge R2(k,c,j,b)))$$

*where $Path(i,k,j)$ indicates that there is a path of arcs of length two connecting $i$ and $j$ which goes through $k$.*

Node consistency is easily enforced by the operation $L_i = L_i \cap \{x | R1(i,x)\}$, requiring $O(nl)$ time (where $n$ is the number of variables and $l$ is the maximum domain size). Arc consistency is enforced by ensuring that every label for a node is supported by at least one label for each node with which it shares a binary constraint (Mackworth, 1977; Mackworth & Freuder, 1985; Mohr & Henderson, 1986). The arc consistency algorithm AC-4 (Mohr & Henderson, 1986) has an worst-case running time of $\Theta(el^2)$ (where $e$ is the number of constraint arcs). AC-3 (Mackworth & Freuder, 1985) often performs better than AC-4 in practice, though it has a slower running time in the worst case. AC-6 (Bessière, 1994) has the same worst-case running time as AC-4 and is faster than AC-3 and AC-4 in practice. Path consistency ensures that any pair of labelings $(i,a) - (j,b)$ allowed by the $(i,j)$ arc directly are also allowed by all arc paths from $i$ to $j$. Montanari has proven that to ensure path consistency for a complete graph, it suffices to check every arc path of length two (Montanari, 1974). The path consistency algorithm PC-4 (Han & Lee, 1988) has a worst-case running time of $O(n^3 l^3)$ time (where $n$ is the number of variables in the CSP).

## 1.2 The Multiply Segmented Constraint Satisfaction Problem

There are many types of problems which can be solved by using CSP in a more or less direct fashion. There are also problems which might benefit from the CSP approach, but which are difficult to represent with a single CSP. This is the class of problems our paper addresses. For example, suppose the map represented in Figure 1 is scanned by a noisy computer vision system, with a resulting uncertainty as to whether the line between regions 1 and 2 is really a border or an artifact of the noise. This situation would yield two CSP problems as depicted in Figure 2. A brute-force approach would be to solve both of the problems, which would be reasonable for scenes containing only a few ambiguous borders. However, as the number of ambiguous borders increases, the number of CSP networks would grow in a combinatorially explosive fashion. In the case of ambiguous segmentation, it can be more efficient to merge the constraint networks into a single network which would compactly represent all of the instances simultaneously, as shown in Figure 3. Notice that the CSP instances are combined into a directed acyclic graph where the paths through the DAG from start to end correspond to those CSPs that were combined. In this paper, we develop an extension to CSP called MUSE CSP (*MU*ltiply *SE*gmented *C*onstraint *S*atisfaction *P*roblem), which represents multiple instances of a CSP problem as a DAG.

If there are multiple, similar instances of a CSP, then separately applying constraints for each instance can result in much duplicated work. To avoid this duplication, we have provided a way to combine the multiple instances of CSP into a MUSE CSP, and we have

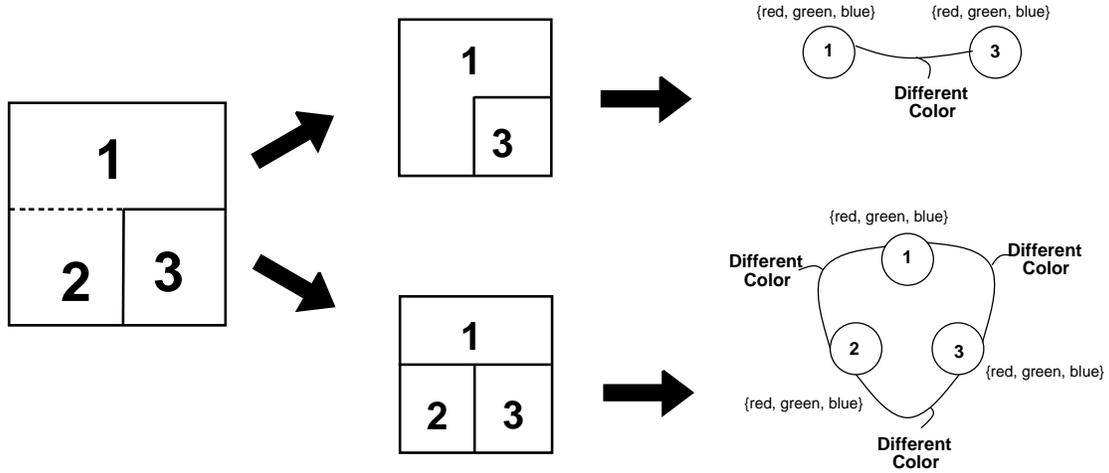

Figure 2: An ambiguous map yields two CSP problems.

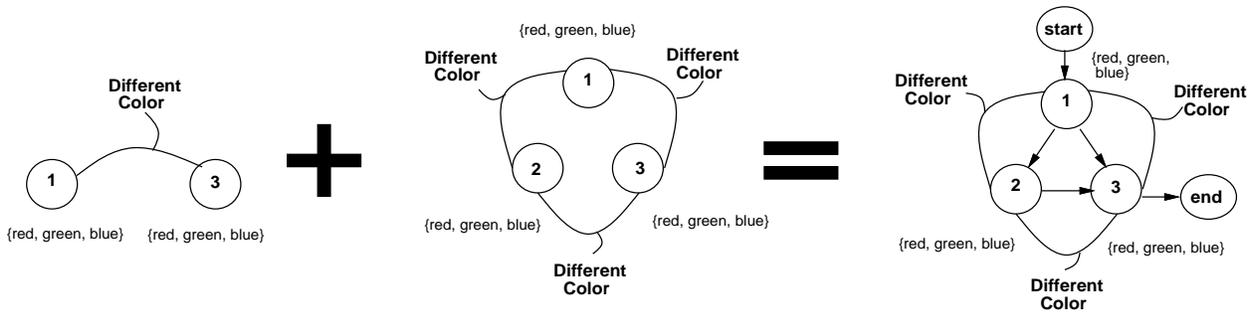

Figure 3: How the two CSP problems of Figure 2 can be captured by a single instance of MUSE CSP. The directed edges form a DAG such that the directed paths through the DAG correspond to instances of those CSPs that were combined.

developed the concepts of MUSE node consistency, MUSE arc consistency, and MUSE path consistency. Formally, we define MUSE CSP as follows:

**Definition 5** (MUSE CSP)
$N = \{i, j, \ldots\}$ *is the set of* nodes *(or variables), with* $|N| = n$,
$\Sigma \subseteq 2^N$ *is a set of* segments *with* $|\Sigma| = s$,
$L = \{a, b, \ldots\}$ *is the set of* labels, *with* $|L| = l$,
$L_i = \{a | a \in L \text{ and } (i, a) \text{ is admissible in at least one segment}\}$,
*R1 is a unary constraint, and* $(i, a)$ *is admissible if* $R1(i, a)$,
*R2 is a binary constraint,* $(i, a) - (j, b)$ *is admissible if* $R2(i, a, j, b)$.

The segments in $\Sigma$ are the different sets of nodes representing CSP instances which are combined to form a MUSE CSP. A solution to a MUSE CSP is defined to be a solution to any one of its segments:

**Definition 6** (Solution to a MUSE CSP) *A solution to a MUSE CSP is an assignment $\alpha$ to a segment $\sigma = \{i_1, \ldots, i_p\}$ such that $\sigma \in \Sigma$ and $\alpha \in L_{i_1} \times \cdots \times L_{i_p}$ such that $R1(i_x, \alpha(i_x))$ holds for every node $i_x \in \sigma$, and $R2(i_x, \alpha(i_x), i_y, \alpha(i_y))$ holds for every pair of nodes $i_x, i_y \in \sigma$, such that $i_x \neq i_y$.*

Depending on the application, the solution for a MUSE CSP could also be the set of all consistent labels for a single path through the MUSE CSP, a single set of labels for each of the paths (or CSPs), or all compatible sets of labels for each of the paths.

A MUSE CSP can be solved with a modified backtracking algorithm which finds a consistent label assignment for a segment. However, when the constraints are sufficiently tight, the search space can be pruned by enforcing local consistency conditions, such as node, arc, and path consistency. To gain the efficiency resulting from enforcing local consistency conditions before backtracking, node, arc, and path consistency must be modified for MUSE CSP. The definitions for MUSE CSP node consistency, arc consistency, and path consistency appear in Definitions 7, 8, and 9, respectively.

**Definition 7** (MUSE Node Consistency) *An instance of MUSE CSP is said to be* node consistent *if and only if each node's domain $L_i$ contains only labels for which the unary constraint R1 holds, i.e.:*

$$\forall i \in N : \forall a \in L_i : R1(i, a)$$

**Definition 8** (MUSE Arc Consistency) *An instance of MUSE CSP is said to be* MUSE arc consistent *if and only if for every label $a$ in each domain $L_i$ there is at least one segment $\sigma$ whose nodes' domains contain at least one label $b$ for which the binary constraint R2 holds, i.e.:*

$$\forall i \in N : \forall a \in L_i : \exists \sigma \in \Sigma : i \in \sigma \land \forall j \in \sigma : j \neq i \Rightarrow \exists b \in L_j : R2(i, a, j, b)$$

**Definition 9** (MUSE Path Consistency) *An instance of MUSE CSP is said to be* path consistent *if and only if:*

$$\forall i, j \in N : i \neq j \Rightarrow (\forall a \in L_i : \forall b \in L_j : \exists \sigma \in \Sigma : i, j \in \sigma \land \forall k \in \sigma : k \neq i \land k \neq j \land Path(i, k, j) \Rightarrow$$
$$(R2(i, a, j, b) \Rightarrow \exists c \in L_k : R2(i, a, k, c) \land R2(k, c, j, b)))$$

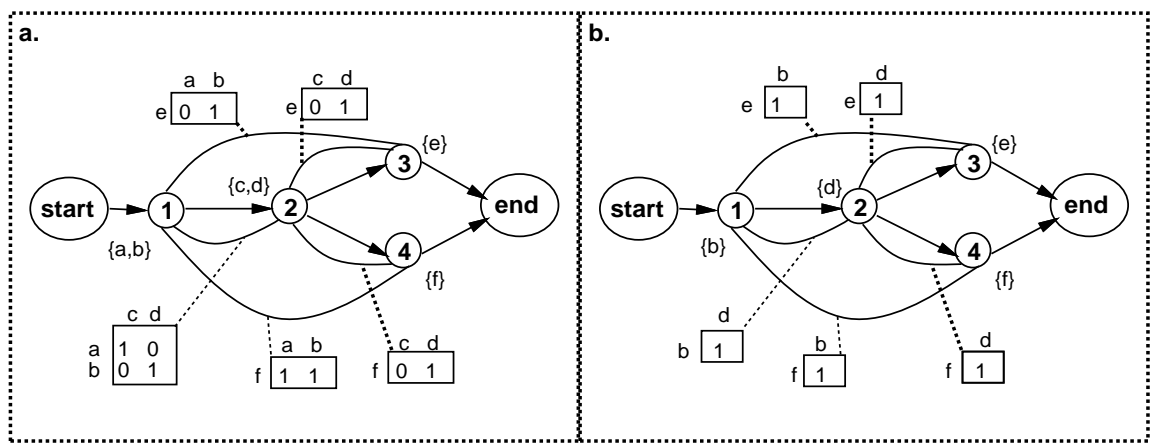

Figure 4: a. A MUSE CSP before MUSE arc consistency is achieved; b. A MUSE CSP after MUSE arc consistency is achieved.

A MUSE CSP is node consistent if all of its segments are node consistent. Unfortunately, MUSE CSP arc consistency requires more attention. When enforcing arc consistency in a CSP, a label $a \in L_i$ can be eliminated from node $i$ whenever any other domain $L_j$ has no labels which together with $a$ satisfy the binary constraints. However, in a MUSE CSP, before a label can be eliminated from a node, it must be unsupported by the arcs of every segment in which it appears, as required by the definition of MUSE arc consistency shown in Definition 8. Notice that Definition 8 reduces to Definition 3 when the number of segments is one.

To demonstrate how MUSE arc consistency applies to a MUSE CSP, consider the MUSE CSP in Figure 4a. Notice that label $c \in L_2$ is not supported by any of the labels in $L_3$ and $L_4$, but does receive support from the labels in $L_1$. Should this label be considered to be MUSE arc consistent? The answer is no because node 2 is only a member of paths through the DAG which contain node 3 or node 4, and neither of them support the label $c$. Because there is no segment such that all of its nodes have some label which supports $c$, $c$ should be eliminated from $L_2$. Once $c$ is eliminated from $L_2$, $a$ will also be eliminated from $L_1$. This is because the elimination of $c$ from $L_2$ causes $a$ to loose the support of node 2. Since node 2 is a member of every path, no other segment provides support for $a$. The MUSE arc consistent DAG is depicted in Figure 4b. Note that MUSE arc consistency does not ensure that the individual segments are arc consistent as CSPs. For example, Figure 5 is MUSE arc consistent even though its segments are not CSP arc consistent. This is because $c$ receives arc support (which is a very local computation) from the arcs of at least one of the paths. We cannot ensure that the values that support a label are themselves mutually consistent by considering MUSE arc consistency alone. For this case, MUSE path consistency together with MUSE arc consistency would be needed to eliminate the illegal labels $c$ and $a$.

When enforcing path consistency in a CSP, $R2(i, a, j, b)$ becomes false if, for any third node $k$, there is no label $c \in L_k$ such that $R2(i, a, k, c)$ and $R2(k, c, j, b)$ are true. In MUSE CSP, if a binary constraint becomes path inconsistent in one segment, it could still be allowed in another. Therefore, the definition of MUSE path consistency is modified as shown in Definition 9.

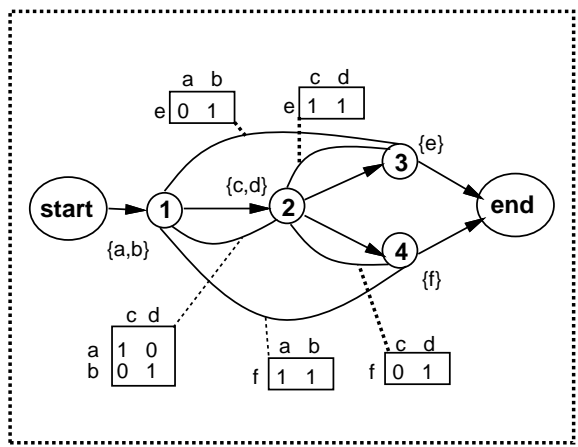

Figure 5: A MUSE CSP which is MUSE arc consistent, but not arc consistent for each segment.

Enforcement of MUSE arc and path consistency requires modification of the traditional CSP algorithms. These algorithms will be described after we introduce several applications for which MUSE CSP has proven useful.

## 2. MUSE CSP and Constraint-based Parsing

It is desirable to represent a MUSE CSP as a directed acyclic graph (DAG) where the directed paths through the DAG correspond to instances of CSP problems. It is often easy to determine which variables should be shared and how to construct the DAG. The application presented in this section is one for which MUSE CSP is useful. A parsing problem is naturally represented as a DAG because of the presence of ambiguity. In many cases, the same word can have multiple parts of speech; it is convenient to represent those words as nodes in a MUSE CSP. In speech recognition systems, the identification of the correct words in a sentence can be improved by using syntactic constraints. However, a word recognition algorithm often produces a lattice of word candidates. Clearly, individually parsing each of the sentences in a lattice can be inefficient.

### 2.1 Parsing with Constraint Dependency Grammar

Maruyama developed a new grammar called Constraint Dependency Grammar (CDG) (Maruyama, 1990a, 1990b, 1990c). He then showed how CDG parsing can be cast as a CSP with a finite domain, so constraints can be used to rule out ungrammatical sentences. A CDG is a four-tuple, $\langle \Sigma, R, L, C \rangle$, where:

$\Sigma$ = a finite set of preterminal symbols, or lexical categories.
$R$ = a finite set of uniquely named roles (or role-ids) = $\{r_1, \ldots, r_p\}$.
$L$ = a finite set of labels = $\{l_1, \ldots, l_q\}$.
$C$ = a finite set of constraints that an assignment **A** must satisfy.

A sentence $s = w_1 w_2 w_3 \ldots w_n \in \Sigma^*$ is a string of length $n$. For each word $w_i \in \Sigma$ of a sentence $s$, we must keep track of $p$ different roles (or variables). A role is a variable which takes on *role values* of the form $<l, m>$, where $l \in L$ and $m \in \{nil, 1, 2, \ldots n\}$. Role values are denoted in examples as *label-modifiee*. In parsing, each label in $L$ indicates a different syntactic function. The value of $m$ in the role value $<l, m>$, when assigned to a particular role of $w_i$, specifies the position of the word that $w_i$ is modifying when it takes on the function specified by the label, $l$ (e.g., subj-3 indicates that the word with that label is a subject when it modifies the third word in the sentence). The sentence $s$ is said to be *generated* by the grammar $G$ if there exists an assignment **A** which maps a role value to each of the $n*p$ roles for $s$ such that the constraint set $C$ (described in the next paragraph) is satisfied.

A *constraint set* is a logical formula of the form: $\forall x_1, x_2, \ldots, x_a$ (and $P_1 P_2 \ldots P_m$), where each $x_i$ ranges over all of the role values in each of the roles for each word of $s$. Each subformula $P_i$ in $C$ must be of the form: (if *Antecedent Consequent*), where *Antecedent* and *Consequent* are predicates or predicates joined by the logical connectives. Below are the basic components used to express constraints.

- **Variables:** $x_1, x_2, \ldots x_a$ ($a = 2$ in (Maruyama, 1990a)).

- **Constants:** elements and subsets of $\Sigma \cup L \cup R \cup \{nil, 1, 2, \ldots, n\}$, where $n$ corresponds to the number of words in a sentence.

- **Functions:**

  **(pos x)** returns the position of the word for role value $x$.

  **(rid x)** returns the role-id for role value $x$.

  **(lab x)** returns the label for role value $x$.

  **(mod x)** returns the position of the modifiee for role value $x$.

  **(cat y)** returns the category (i.e., the element in $\Sigma$) for the word at position $y$.

- **Predicates:** $=, >, <^1$.

- **Logical Connectives:** and, or, not.

A subformula $P_i$ is called a *unary constraint* if it contains one variable and a *binary constraint* if it contains two. A CDG grammar has two associated parameters, *degree* and *arity*. The degree of a grammar G is the number of roles. The arity of the grammar, $a$, corresponds to the maximum number of variables in the subformulas of $C$.

Consider the example grammar, $G_1$, which is defined using the following four-tuple: $\langle \Sigma_1 = \{\text{det}, \text{noun}, \text{verb}\}, R_1 = \{\text{governor}\}, L_1 = \{\text{det}, \text{root}, \text{subj}\}, C_1$ (see constraints in Figure 6)$\rangle$. $G_1$ has a degree of one and an arity of two. To illustrate the process of parsing with constraint satisfaction, Figure 6 shows the steps for parsing the sentence *The dog eats*. To simplify the presentation of this example, the grammar uses a single role, the *governor* role, which is denoted as **G** in the constraint network in Figure 6. The governor role indicates the function a word fills in a sentence when it is governed by its head word. A word is called the head of a phrase when it forms the basis of the phrase (e.g., the verb is the head of the sentence). In useful grammars, we would also include several needs roles (e.g, need1, need2) to make certain that a head word has all of the constituents it needs to be complete (e.g., a singular count noun needs a determiner to be a complete noun phrase). To determine whether the sentence, *The dog eats*, is generated by the grammar, the CDG parser must be able to assign at least one role value to each of the $n * p$ roles that satisfies the grammar constraints ($n = 3$ is sentence length, and $p = 1$ is the number of roles). Because the values for a role are selected from the finite set $L_1 \times \{$nil, 1, 2, 3$\}$, CDG parsing can be viewed as a constraint satisfaction problem over a finite domain. Therefore, constraint satisfaction can be used to determine the possible parses of this sentence.

---

1. Note that $1 > nil$ or $1 < nil$ is false, because nil is not an integer. For MUSE networks, we relate position intervals using $<, >$, and $=$.

Initially, for each word, all possible role values are assigned to the governor role. We assume that a word must either modify another word (other than itself) or modify no word ($m$=nil). Nothing is gained in CDG by having a word modify itself. Next the unary constraints are applied to all of the role values in the constraint network. A role value is incompatible with a unary constraint if and only if it satisfies the antecedent, but not the consequent. Notice in Figure 6 that all the role values associated with the governor role of the first word (*the*) satisfy the antecedent of the first unary constraint, but det-nil, subj-nil, subj-2, subj-3, root-nil, root-2, and root-3 do not satisfy the consequent, and so they are incompatible with the constraint. When a role value violates a unary constraint, node consistency eliminates those role values from their role because they can never participate in a parse for the sentence. After all unary constraints are applied to the top constraint network in Figure 6, the second network is produced.

Next, binary constraints are applied. Binary constraints determine which pairs of role values can legally coexist. To keep track of pairs of role values, *arcs* are constructed connecting each role to all other roles in the network, and each arc has an associated *arc matrix*, whose row and column indices are the role values associated with the two roles it connects. The entries in an arc matrix can either be **1** (indicating that the two role values indexing the entry are compatible) or **0** (indicating that the role values cannot simultaneously exist). Initially, all entries in each matrix are set to **1**, indicating that the pair of role values indexing the entry are initially compatible (because no constraints have been applied). In our example, the single binary constraint (shown in Figure 6) is applied to the pairs of role values indexing the entries in the matrices. For example, when x=det-3 for *the* and y=root-nil for *eats*, the consequent of the binary constraint fails; hence, the role values are incompatible. This is indicated by replacing the entry of **1** with **0**.

Following the binary constraints, the roles of the constraint network can still contain role values which are incompatible with the parse for the sentence. Role values that are not supported by the binary constraints can be eliminated by achieving arc consistency. For example, det-3 for *the* is not supported by the remaining role value for *eats* and is thus deleted from the role.

After arc consistency, the example sentence has a single parse because there is only one value per role in the sentence. A parse for a sentence consists of an assignment of role values

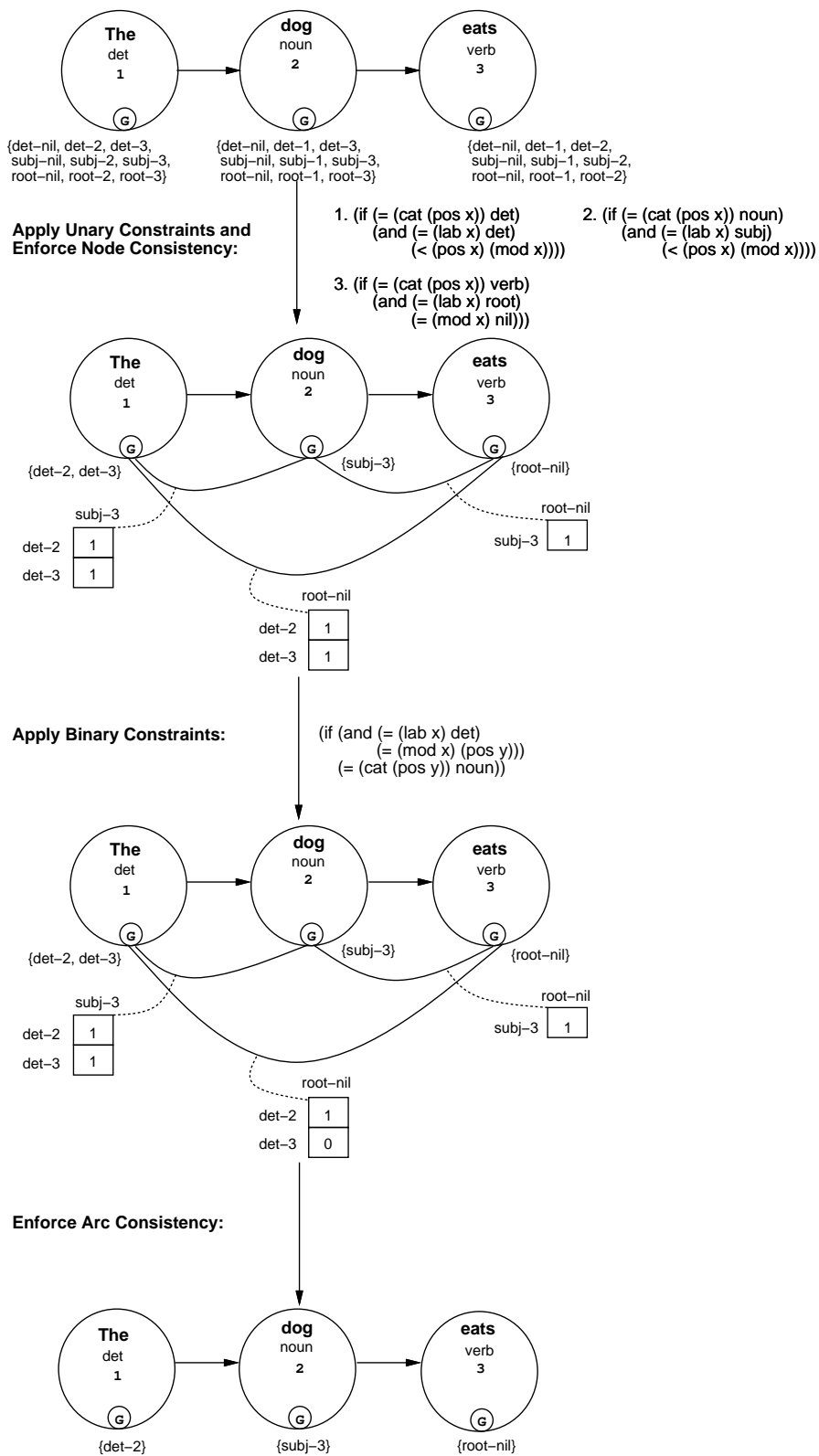

Figure 6: Using constraints to parse the sentence: *The dog eats.*

to roles such that the unary and binary constraints are satisfied for that assignment. In general, there can be more than one parse for a sentence; hence, there can be more than one assignment of values to the roles of the sentence. Note that the assignment for the example sentence is:

| pos | word | cat | governor role's value |
|-----|------|-----|----------------------|
| **1** | the | det | det-2 |
| **2** | dog | noun | subj-3 |
| **3** | eats | verb | root-nil |

If there is only one possible sentence such that the part of speech of each of the words is known in advance, then the parsing problem can be cast as a CSP. However, for the ambiguity present in written and spoken sentences to be handled uniformly requires the use of MUSE CSP.

## 2.2 Processing Lexically Ambiguous Sentences with CDG

One shortcoming of Maruyama's constraint-based parser is that it requires a word to have a single part of speech; however, many words in the English language have more than one lexical category. This assumption is captured in the way that Maruyama writes constraints involving category information; the category is determined based on the position of the word in the sentence. However, even in our simple example, the word *dog* could have been either a noun or a verb prior to the propagation of syntactic constraints. Since parsing can be used to lexically disambiguate the sentence, ideally, the parsing algorithm should not require that the part of speech for the words be known prior to parsing.

Lexically ambiguous words can easily be accommodated by creating a CSP for each possible combination of lexical categories; however, this would be combinatorially explosive. In contrast, using a MUSE CSP, we can create a separate word node for each legal part of speech of a word, sharing those words that are not ambiguous across all segments. Since position does not uniquely define the category of a word, we must allow category information to be accessed through the role value rather than the position of the word in the sentence (i.e., use `(cat x)` rather than `(cat (pos x))`). Once we associate category information with a role value, we could instead create role values for each lexical category for a word and store all of the values in a single word node. However, this representation is not as convenient as the MUSE CSP representation for the problem. In the lexically augmented CSP, when there is more than one role per word (this is usually the case), the role values associated with one lexical category for one role cannot support the role values associated with another lexical category in another role for the same word. Additional constraints must be propagated to enforce this requirement. The MUSE CSP representation does not suffer from this problem. By using a separate node for each part of speech, the MUSE CSP directly represents the independence of the alternative lexical categories for a given word. The space requirements for the arc matrices in the MUSE representation is lower than for the lexicalized CSP as there is no arc between the roles for the different lexical categories for a word in the MUSE representation. Note that MUSE arc consistency is equivalent to performing arc consistency on the lexically augmented CSP (after the additional constraints are propagated)[2]. Most importantly, MUSE CSP can represent lattices that cannot be combined into a single CSP.

The technique of creating separate nodes for different instances of a word can also be used to handle feature analysis (like number and person) in parsing (Harper & Helzerman, 1995b). Since some words have multiple feature values, it is often more efficient to create a single node with a set of feature values, apply syntactic constraints, and then split the node into a set of nodes with a single feature value prior to applying the constraints pertaining to the feature type. Node splitting can also be used to support the use of context-specific constraints (Harper & Helzerman, 1995b).

## 2.3 Lattice Example

Much of the motivation for extending CSP comes from our work in spoken language parsing (Harper & Helzerman, 1995a; Harper, Jamieson, Zoltowski, & Helzerman, 1992; Zoltowski, Harper, Jamieson, & Helzerman, 1992). The output of a hidden-Markov-model-based speech recognizer can be thought of as a lattice of word candidates. Unfortunately, a lattice contains many word candidates that can never appear in a sentence covering the duration of a speech utterance. By converting the lattice to a word graph, many word candidates in the lattice can be eliminated. Figure 7 depicts a word graph constructed from a simple lattice. Notice that the word *tour* can be eliminated when the word graph is constructed. In order to accommodate words that occur over time intervals that may overlap, each word's position in the lattice is now represented as a tuple $(b, e)$ such that $b < e$. The positional relations defined for constraints are easily modified to operate on tuples (Harper & Helzerman, 1995a).

After construction, the word graph often contains spurious sentence hypotheses which can be pruned by using a variety of constraints (e.g., syntactic, semantic, etc.). We can apply constraints to individual sentences to rule out those that are ungrammatical; however, individually processing each sentence hypothesis is inefficient since many have a high degree of similarity. If the spoken language parsing problem is structured as a MUSE CSP problem, then the constraints used to parse individual sentences would be applied to the word graph of sentence hypotheses, eliminating from further consideration many hypotheses which are ungrammatical.

We have developed a MUSE CSP constraint-based parser, PARSEC (Harper & Helzerman, 1995a, 1995b; Harper et al., 1992; Zoltowski et al., 1992), which is capable of parsing word graphs containing multiple sentences produced by a speech recognition module. We have developed syntactic and semantic constraints for parsing single sentences, which when applied to a word graph, eliminate those hypotheses that are syntactically or semantically incorrect. The MUSE CSP used by our parser can be thought of as a parse forest which is pruned by using constraints. By applying constraints from a wide variety of knowledge sources, the parser prunes the composite structure of many of the role values associated with a role, as well as word nodes with no remaining role values. Several experiments (Harper et al., 1992; Zoltowski et al., 1992) have considered how effective syntactic and semantic constraints are at pruning word nodes that can appear in no sentence hypothesis. For our work in speech processing, the MUSE arc consistency algorithm is very effective at pruning the role values from the composite structure that can never appear in a parse for a sentence (i.e., an individual CSP). Constraints are usually tight enough that MUSE arc consistency eliminates role values that do not participate in at least one parse for the represented sentences.

---

2. As a simple demonstration, consider merging nodes 3 and 4 from Figure 5 into a single node such that the value $e$ and $f$ keep track of the fact that they have type 3 and 4, respectively. Under these circumstances, CSP arc consistency will give the same results as MUSE CSP arc consistency; even though $a$ and $c$ appear in no solutions, they are not eliminated. Note that this example uses only one role per node.

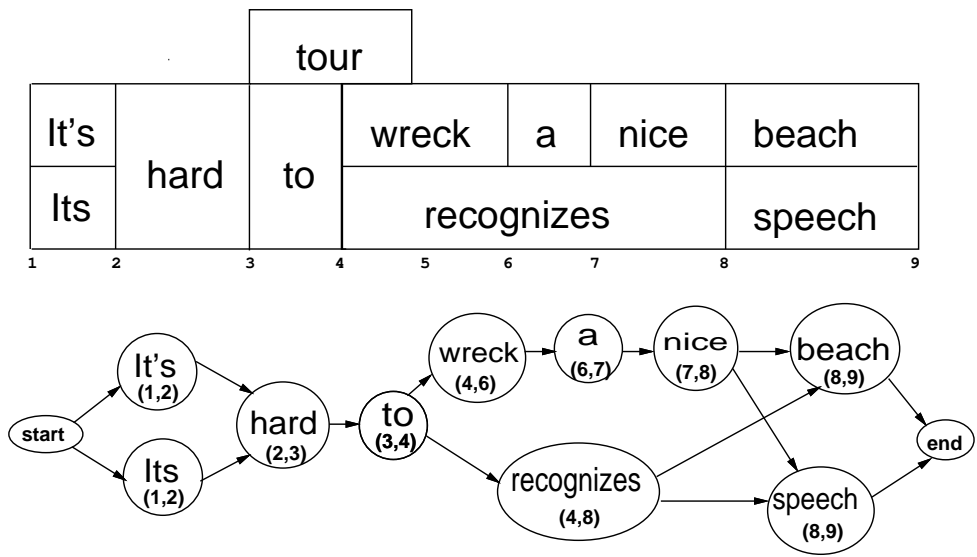

Figure 7: Multiple sentence hypotheses can be parsed simultaneously by applying constraints over a word graph rather than individual sentences extracted from a lattice.

MUSE CSP is a useful way to process multiple sentences because the arc consistency algorithm is effective at eliminating role values that cannot appear in sentence parses. Several factors contribute to the effectiveness of the arc consistency algorithm for this problem. First, the syntactic constraints are fairly tight constraints. Second, the role values contain some segmental information that constrain the problem. Consider the word graph in Figure 8. The value s-(3,4) associated with the role marked **N** for the word *are* cannot support any of the values for the role marked **G** for the word *dogs* at position (3,5), because it is not legal in a segment involving position (3,5). In the figure, we mark those entries where a value associated with one role is segmentally incompatible with the values of another with an N. These entries are equivalent to 0. Third, many times constraints create symmetric dependencies between words in the sentence. For example, one constraint might indicate that a verb needs a subject to its left, and another that a subject must be governed by a verb to its right.

## 2.4 A Demonstration of the Utility of MUSE CSP Parsing

To demonstrate the utility of MUSE CSP for simultaneously parsing multiple CSP instances, consider the problem of determining which strings of length $3n$ consisting of $a$'s, $b$'s, and $c$'s are in the language $a^n b^n c^n$. For the value of $n = 3$, this problem can be represented as the single MUSE CSP problem shown in Figure 9 (the roles and role values are not depicted to simplify the figure). We have devised constraints for this language (see Figure 10) which eliminate all role values for all sentences not in the language as well as all ungrammatical role values for a sentence in the language. When these constraints are applied followed by MUSE arc consistency to a lattice like that in Figure 9 with a length divisible by three, then only the grammatical sentence will remain with a single parse. For lattices containing only sentences with lengths that are not divisible by three, all role values are eliminated by MUSE arc consistency (there is no grammatical sentence). Hence, there is no search required to extract a parse if there is one. For the $n = 3$ case of Figure 9, the parse appears in Figure 11. A single parse will result regardless of the $n$ chosen. Note that the modifiees for the role values in the parse are used to ensure that for each $a$, there is a corresponding $c$; for each $b$, there is a corresponding $a$; and for each $c$, there is a corresponding $b$. Figure 12 examines the time needed to extract a parse for sentences in the language $a^n b^n c^n$ from MUSE CSPs representing all strings of length $3n$, $1 \le n \le 7$, containing $a$, $b$, and $c$. The time to perform MUSE AC-1 and extract the solution is compared to the time to extract the solution without any preprocessing. The time to perform MUSE AC-1 and extract the parse is stable as sentence length grows, but the time to extract a parse grows quickly for sentence lengths greater than 15 when MUSE arc consistency is not used.

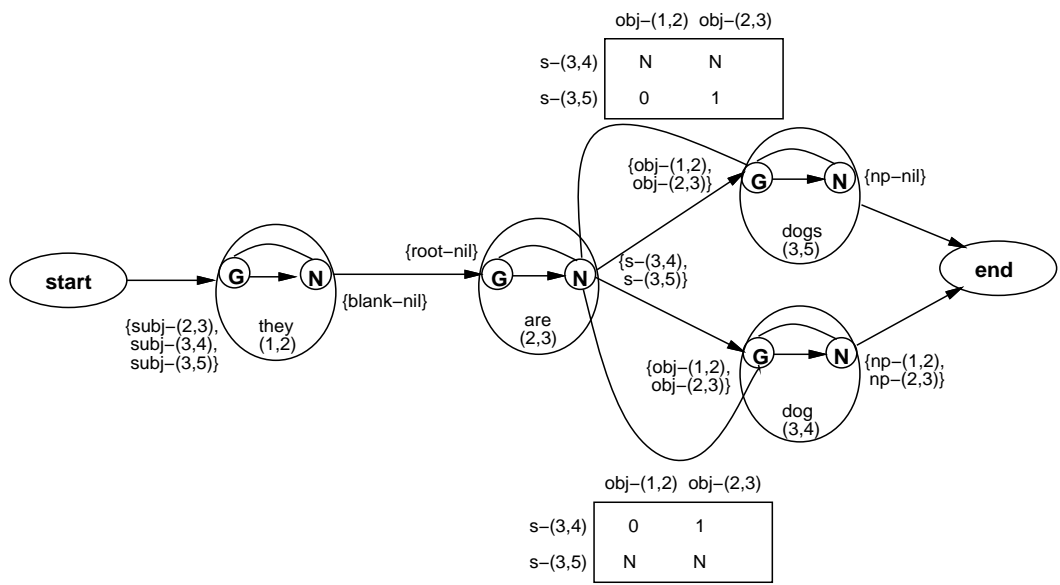

Figure 8: In parsing word graphs, some of the values assigned to roles contain segmental information which make them incompatible with the values associated with some of the other roles. For example, s-(3,4) cannot support any of the values associated with the **G** or **N** roles of the word *dogs*.

The previous example involves a grammar where there can only be one parse for a single sentence in the lattice; however, it is a simple matter to provide similar demonstrations for

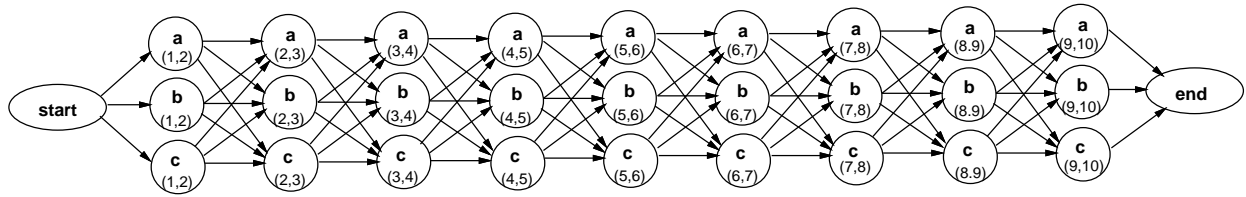

Figure 9: A single MUSE CSP can simultaneously test all possible orderings of $a$'s, $b$'s, and $c$'s for membership in the language $a^n b^n c^n$, $n = 3$.

$\Sigma_2 = \{a, b, c\}$
$R_2 = \{governor\}$
$L_2 = \{a, b, c\}$
$C_2 = $ see below:

```
                       ; 3 Unary Constraints

(if (and (= (cat x) a)                 (if (and (= (cat x) b)
         (= (rid x) governor))                  (= (rid x) governor))
    (and (= (lab x) a)                     (and (= (lab x) b)
         (> (mod x) (pos x))))                  (< (mod x) (pos x))))

(if (and (= (cat x) c)
         (= (rid x) governor))
    (and (= (lab x) c)
         (< (mod x) (pos x))))

                       ; 8 Binary Constraints

(if (and (= (lab x) a)                 (if (and (= (lab x) b)
         (or (= (lab y) b)                      (= (lab y) c))
             (= (lab y) c)))              (< (pos x) (pos y)))
    (< (pos x) (pos y)))

(if (and (= (lab x) a)                 (if (and (= (lab x) a)
         (= (lab y) a)                          (= (mod x) (pos y))
         (> (pos x) (pos y)))                   (= (rid y) governor))
    (< (mod x) (mod y)))                   (= (lab y) c))

(if (and (= (lab x) b)                 (if (and (= (lab x) b)
         (= (lab y) b)                          (= (mod x) (pos y))
         (> (pos x) (pos y)))                   (= (rid y) governor))
    (< (mod x) (mod y)))                   (= (lab y) a))

(if (and (= (lab x) c)                 (if (and (= (lab x) c)
         (= (lab y) c)                          (= (mod x) (pos y))
         (> (pos x) (pos y)))                   (= (rid y) governor))
    (< (mod x) (mod y)))                   (= (lab y) b))
```

Figure 10: $G_2 = \langle \Sigma_2, R_2, L_2, C_2 \rangle$ accepts the language $a^n b^n c^n$, $n \geq 0$.

253

| pos | cat | governor role's value |
|-----|-----|------------------------|
| **(1,2)** | a | a-(9,10) |
| **(2,3)** | a | a-(8,9) |
| **(3,4)** | a | a-(7,8) |
| **(4,5)** | b | b-(3,4) |
| **(5,6)** | b | b-(2,3) |
| **(6,7)** | b | b-(1,2) |
| **(7,8)** | c | c-(6,7) |
| **(8,9)** | c | c-(5,6) |
| **(9,10)** | c | c-(4,5) |

Figure 11: The single parse remaining in the network depicted in Figure 9 after the applying the constraints in $G_2$ and enforcing MUSE arc consistency.

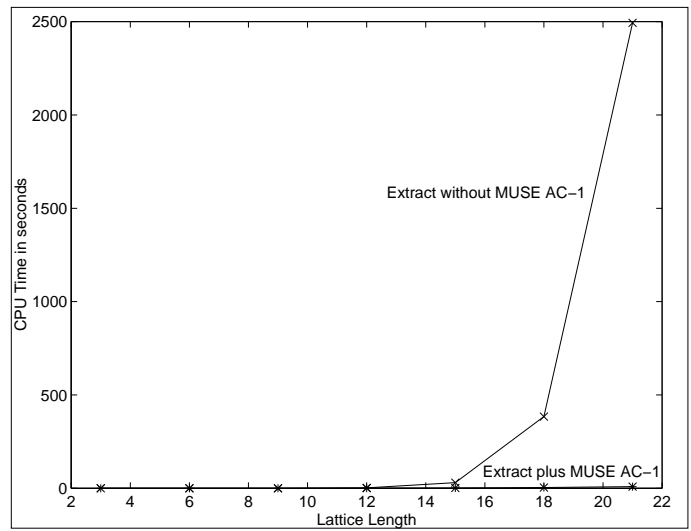

Figure 12: This graph depicts the time to extract the parse for the language $a^n b^n c^n$ from a MUSE CSP representing all sentences of length $3n$, where $n$ varies from 1 to 7. The time to extract the parse without MUSE arc consistency is compared to the time to perform MUSE AC-1 and extract the parse.

```
Σ₃ = {a, b, c}
R₃ = {governor}
L₃ = {w1, w2}
C₃ = see below:

                    ; 2 Unary Constraints

(if (= (lab x) w1)                      (if (= (lab x) w2)
    (< (pos x) (mod y)))                    (> (pos x) (mod y)))

                    ; 6 Binary Constraints

(if (and (= (lab x) w1)                 (if (and (= (lab x) w1)
         (= (lab y) w2))                         (= (lab y) w2))
    (< (pos x) (pos y)))                    (> (mod x) (mod y)))

(if (and (= (lab x) w1)                 (if (and (= (lab x) w2)
         (= (lab y) w1)                          (= (lab y) w2)
         (> (pos x) (pos y)))                    (> (pos x) (pos y)))
    (> (mod x) (mod y)))                    (< (mod x) (mod y)))

(if (and (= (lab x) w1)                 (if (and (= (lab x) w2)
         (= (mod x) (pos y)))                    (= (mod x) (pos y)))
    (and (= (lab y) w2)                     (= (lab y) w1))
         (= (cat x) (cat y))))
```

Figure 13: $G_3 = \langle \Sigma_3, R_3, L_3, C_3 \rangle$ accepts the language $ww$.

more complex cases. For example, the constraint grammar shown in Figure 13 can be to parse all possible sentences of a given length in the the language $ww$, such that $w$ is in $\{a, b, c\}^+$. Consider the MUSE CSP in Figure 14 (the roles and role values are not depicted to simplify the figure). After applying the constraints and performing MUSE arc consistency on this MUSE CSP, there are precisely 81 strings that are in $ww$, and their parses are compactly represented in the constraint network. The constraints plus MUSE arc consistency eliminate every value that cannot appear in a parse. For lattices containing odd length sentences, no role values remain after MUSE arc consistency. Figure 15 shows the time needed to extract all of the parses for sentences in the language $ww$ from the MUSE CSPs as we vary the length of $w$ from 1 to 8. The time to perform MUSE AC-1 and extract the parses grows slowly as sentence length increases because the number of parses increases with sentence length; however, it grows more slowly than the time to extract the parses when MUSE arc consistency is not used.

Similar results have also been obtained with grammars used to parse word graphs constructed from spoken sentences in the resource management and ATIS domains (Harper et al., 1992; Zoltowski et al., 1992; Harper & Helzerman, 1995a).

## 3. The MUSE CSP Arc Consistency Algorithm

In this section, we introduce an algorithm, MUSE AC-1, to achieve MUSE CSP arc consistency. Because our algorithm builds upon the AC-4 algorithm (Mohr & Henderson, 1986), we present that algorithm first for comparison purposes.

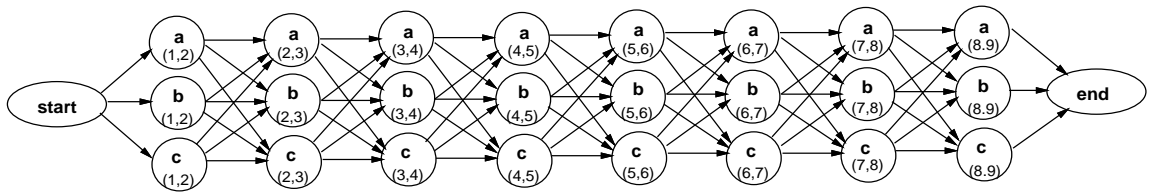

Figure 14: A single MUSE CSP can simultaneously test all possible orderings of $a$'s, $b$'s, and $c$'s for membership in the language $ww$ where $|w| = 4$ .

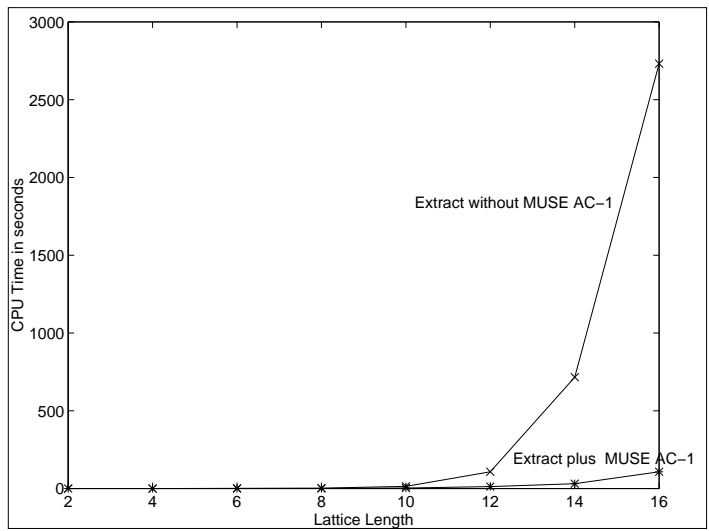

Figure 15: This graph depicts the time to extract all parses for the language $ww$ from a MUSE CSP representing all sentences of length 2 to 16 such that $w \in \{a, b, c\}^+$. The time to extract all parses without MUSE arc consistency is compared to the time to perform MUSE AC-1 and extract all parses.

| Notation | Meaning |
|----------|---------|
| $(i, j)$ | An ordered pair of nodes. |
| $E$ | All node pairs $(i, j)$. If $(i, j) \in E$, then $(j, i) \in E$. |
| $(i, a)$ | An ordered pair of node $i$ and label $a \in L_i$. |
| $L_i$ | $\{a \mid a \in L$ and $(i, a)$ is permitted by the constraints (i.e., admissible)$\}$ |
| $R2(i, a, j, b)$ | $R2(i, a, j, b) = 1$ indicates the admissibility $a \in L_i$ and $b \in L_j$ given binary constraints. |
| $\text{Counter}[(i, j), a]$ | The number of labels in $L_j$ which are compatible with $a \in L_i$. |
| $S[i, a]$ | $(j, b) \in S[i, a]$ means that $a \in L_i$ and $b \in L_j$ are simultaneously admissible. This implies that $a$ supports $b$. |
| $M[i, a]$ | $M[i, a] = 1$ indicates that the label $a$ is not admissible for (and has already been eliminated from) node $i$. |
| $List$ | A queue of arc support to be deleted. |

Figure 16: Data structures and notation for the arc consistency algorithm, AC-4.

## 3.1 CSP Arc Consistency: AC-4

AC-4 builds and maintains several data structures, described in Figure 16, to allow it to efficiently achieve arc consistency in a CSP. Note that we have modified the notation slightly to eliminate subscripts (which become quite cumbersome for the path consistency algorithm). Figure 17 shows the code for initializing the data structures, and Figure 18 contains the algorithm for eliminating inconsistent labels from the domains. This algorithm requires $\Theta(el^2)$ time, where $e$ is the number of constraint arcs, and $l$ is the domain size (Mohr & Henderson, 1986).

In AC-4, if the label $a \in L_i$ is compatible with $b \in L_j$, then $a$ *supports* $b$ (and vice versa). To keep track of how much support each label $a$ has, the number of labels in $L_j$ which are compatible with $a$ in $L_i$ are counted and the total stored in $\text{Counter}[(i, j), a]$ by the algorithm in Figure 17. If any $\text{Counter}[(i, j), a]$ is zero, then $a$ is removed from $L_i$ (because it cannot appear in any solution), the ordered pair $(i, a)$ is placed on the $List$, and $M[i, a]$ is set to 1 (to avoid removing the element $a$ from $L_i$ more than once). The algorithm must also keep track of which labels that label $a$ supports by using $S[i, a]$, a set of arc and label pairs. For example, $S[i, a] = \{(j, b), (j, c)\}$ means that $a$ in $L_i$ supports $b$ and $c$ in $L_j$. If $a$ is ever removed from $L_i$, then $b$ and $c$ will loose some of their support.

After the preprocessing step in Figure 17, the algorithm in Figure 18 loops until $List$ becomes empty, at which point the CSP is arc consistent. When $(i, a)$ is popped off $List$ by this procedure, for each element $(j, b)$ in $S[i, a]$, $\text{Counter}[(j, i), b]$ is decremented. If $\text{Counter}[(j, i), b]$ becomes zero, $b$ would be removed from $L_j$, $(j, b)$ placed on $List$, and $M[j, b]$ set to 1.

```
1.    List := φ;
2.    for i ∈ N do
3.        for a ∈ Lᵢ do {
4.            S[i, a] := φ;
5.            M[i, a] := 0; }
6.    for (i, j) ∈ E do
7.        for a ∈ Lᵢ do {
8.            Total := 0;
9.            for b ∈ Lⱼ do
10.               if R2(i, a, j, b) then {
11.                   Total := Total+1;
12.                   S[j, b] := S[j, b] ∪ {(i, a)}; }
13.           if Total = 0 then {
14.               Lᵢ := Lᵢ − {a};
15.               List := List ∪ {(i, a)};
16.               M[i, a] := 1; }
17.           Counter[(i, j), a] := Total; }
```

Figure 17: Initialization of the data structures for AC-4.

```
1.    while List ≠ φ do {
2.        pop (i, a) from List;
3.        for (j, b) ∈ S[i, a] do {
4.            Counter[(j, i), b] := Counter[(j, i), b] − 1;
5.            if Counter[(j, i), b] = 0 ∧ M[j, b] = 0 then {
6.                Lⱼ := Lⱼ − {b};
7.                List := List ∪ {(j, b)};
8.                M[j, b] := 1; } } }
```

Figure 18: Eliminating inconsistent labels from the domains in AC-4.

Next, we describe the MUSE arc consistency algorithm for a MUSE CSP, called MUSE AC-1. We purposely keep our notation and presentation of MUSE AC-1 as close as possible to that of AC-4 so that the reader can benefit from the similarity of the two algorithms.

## 3.2 MUSE AC-1

MUSE arc consistency is enforced by removing those labels in each $L_i$ which violate the conditions of Definition 8. MUSE AC-1 builds and maintains several data structures, described in Figure 19, to allow it to efficiently perform this operation. Many of these data structures are borrowed from AC-4, while others exploit DAG representation of the MUSE CSP to determine when values are incompatible in all of the segments. Figure 22 shows the code for initializing the data structures, and Figures 23 and 24 contain the algorithm for eliminating inconsistent labels from the domains.

In MUSE AC-1 as in AC-4, if label $a$ at node $i$ is compatible with label $b$ at node $j$, then $a$ supports $b$. To keep track of how much support each label $a$ has, the number of labels in $L_j$ which are compatible with $a$ in $L_i$ are counted, and the total is stored in Counter$[(i, j), a]$. For CSP arc consistency, if Counter$[(i, j), a]$ is zero, $a$ would be immediately removed from $L_i$, because that would mean that $a$ could never appear in any solution. However, in MUSE arc consistency, this may not be the case, because even though $a$ does not participate in a solution for any of the segments which contain $i$ and $j$, there could be another segment for which $a$ would be perfectly legal. A label cannot become globally inadmissible until it is incompatible with every segment. Hence, in MUSE CSP, if Counter$[(i, j), a]$ is zero, the algorithm simply places $[(i, j), a]$ on *List* and records that fact by setting M$[(i, j), a]$ to 1. By placing $[(i, j), a]$ on *List*, the algorithm is indicating that the segments containing $i$ and $j$ do not support the label $a$.

MUSE AC-1 must also keep track of those labels in $j$ that label $a$ in $L_i$ supports by using S$[(i, j), a]$, a set of node-label pairs. For example, S$[(i, j), a] = \{(j, b), (j, c)\}$ means that $a$ in $L_i$ supports $b$ and $c$ in $L_j$. If $a$ is ever invalid for $L_i$, then $b$ and $c$ will loose some of their support.

Because $\Sigma$ is a DAG, MUSE AC-1 is able to use the properties of the DAG to identify local (and hence efficiently computable) conditions under which labels become globally inadmissible. Segments are defined as paths through the MUSE CSP from start to end. If a value associated with a variable is not supported by any of the variables which precede it or follow it, then there is no way that the value can be used by any segment, so it can be deleted by the arc consistency algorithm. In addition, if a value in a variable's domain is supported by the constraints for values associated with a second variable, but the second variable is preceded or followed by variables that have no values supporting the value, then because a solution involves a path of variables in the MUSE DAG, the value cannot be supported for any segment involving the two variables. These two ideas provide the basis for the remaining data structures used by MUSE AC-1.

Consider Figure 20, which shows the nodes which are adjacent to node $i$ in the DAG. Because every segment in the DAG which contains node $i$ is represented as a directed path in the DAG going through node $i$, either node $j$ or node $k$ must be in every segment containing $i$. Hence, if the label $a$ is to remain in $L_i$, it must be compatible with at least one label in either $L_j$ or $L_k$. Also, because either $n$ or $m$ must be contained in every segment containing

| Notation | Meaning |
|----------|---------|
| $(i, j)$ | An ordered pair of nodes. |
| $E$ | All node pairs $(i, j)$ such that there exists a path of directed edges in $G$ between $i$ and $j$. If $(i, j) \in E$, then $(j, i) \in E$. |
| $(i, a)$ | An ordered pair of node $i$ and label $a \in L_i$. |
| $[(i, j), a]$ | An ordered pair of a node pair $(i, j)$ and a label $a \in L_i$. |
| $L_i$ | $\{a \mid a \in L$ and $(i, a)$ is permitted by the constraints (i.e., admissible)$\}$ |
| $R2(i, a, j, b)$ | $R2(i, a, j, b) = 1$ indicates the admissibility of $a \in L_i$ and $b \in L_j$ given binary constraints. |
| Counter$[(i, j), a]$ | The number of labels in $L_j$ which are compatible with $a \in L_i$. |
| S$[(i, j), a]$ | $(j, b) \in S[(i, j), a]$ means that $a \in L_i$ and $b \in L_j$ are simultaneously admissible. This implies that $a$ supports $b$. |
| M$[(i, j), a]$ | M$[(i, j), a] = 1$ indicates that the label $a$ is not admissible for (and has already been eliminated from) all segments containing $i$ and $j$. |
| *List* | A queue of arc support to be deleted. |
| $G$ | $G$ is the set of node pairs $(i, j)$ such that there exists a directed edge from $i$ to $j$. |
| Next-Edge$_i$ | Next-Edge$_i$ contains all node pairs $(i, j)$ such that there exists a directed edge $(i, j) \in G$. It also contains $(i, \text{end})$ if $i$ is the last node in a segment. |
| Prev-Edge$_i$ | Prev-Edge$_i$ contains all node pairs $(j, i)$ such that there exists a directed edge $(j, i) \in G$. It also contains $(\text{start}, i)$ if $i$ is the first node in a segment. |
| Local-Prev-Support$(i, a)$ | A set of elements $(i, j)$ such that $(j, i) \in$ Prev-Edge$_i$, and if $j \neq \text{start}$, $a$ must be compatible with at least one of $j$'s labels. If Local-Prev-Support$(i, a)$ becomes empty, $a$ in $i$ is no longer admissible. |
| Local-Next-Support$(i, a)$ | A set of elements $(i, j)$ such that $(i, j) \in$ Next-Edge$_i$, and if $j \neq \text{end}$, $a$ must be compatible with at least one of $j$'s labels. If Local-Next-Support$(i, a)$ becomes empty, $a$ in $i$ is no longer admissible. |
| Prev-Support$[(i, j), a]$ | $(i, k) \in$ Prev-Support$[(i, j), a]$ implies that $(k, j) \in$ Prev-Edge$_j$, and if $k \neq \text{start}$, then $a \in L_i$ is compatible with at least one of $j$'s and one of $k$'s labels. If Prev-Support$[(i, j), a]$ becomes empty, then $a$ is no longer admissible in segments containing $i$ and $j$. |
| Next-Support$[(i, j), a]$ | $(i, k) \in$ Next-Support$[(i, j), a]$ implies that $(j, k) \in$ Next-Edge$_j$, and if $k \neq \text{end}$, then $a \in L_i$ is compatible with at least one of $j$'s and one of $k$'s labels. If Next-Support$[(i, j), a]$ becomes empty, then $a$ is no longer admissible in segments containing $i$ and $j$. |

Figure 19: Data structures and notation for MUSE AC-1.

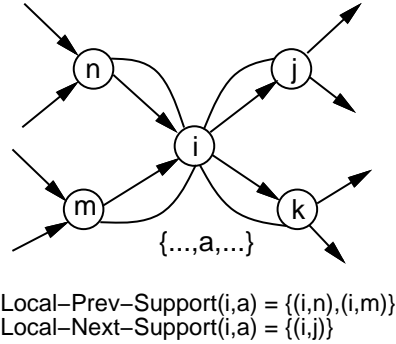

Local-Prev-Support(i,a) = {(i,n),(i,m)}
Local-Next-Support(i,a) = {(i,j)}

Figure 20: Local-Prev-Support and Local-Next-Support for an example DAG. The sets indicate that the label $a$ is allowed for every segment which contains $n$, $m$, and $j$, but is disallowed for every segment which contains $k$. The solid directed lines are members of $G$, and the solid undirected lines represent members of $E$.

$i$, if label $a$ is to remain in $L_i$, it must also be compatible with at least one label in either $L_n$ or $L_m$.

In order to track this dependency, two sets are maintained for each label $a$ at node $i$, Local-Next-Support$(i, a)$ and Local-Prev-Support$(i, a)$. Local-Next-Support$(i, a)$ is a set of ordered node pairs $(i, j)$ such that $(i, j) \in$ Next-Edge$_i$, and if $(i, j) \in E$, there is at least one label $b \in L_j$ which is compatible with $a$. Local-Prev-Support$(i, a)$ is a set of ordered pairs $(i, j)$ such that $(j, i) \in$ Prev-Edge$_i$, and if $(i, j) \in E$, there is at least one label $b \in L_j$ which is compatible with $a$. Dummy ordered pairs are also created to handle cases where a node is at the beginning or end of a network: when (start, $i$) $\in$ Prev-Edge$_i$, $(i, \text{start})$ is added to Local-Prev-Support$(i, a)$, and when $(i, \text{end}) \in$ Next-Edge$_i$, $(i, \text{end})$ is added to Local-Next-Support$(i, a)$. This is to prevent a label from being ruled out because no nodes precede or follow it in the DAG. Whenever one of $i$'s adjacent nodes, $j$, no longer has any labels $b$ in its domain which are compatible with $a$, then $(i, j)$ should be removed from Local-Prev-Support$(i, a)$ or Local-Next-Support$(i, a)$, depending on whether the edge is from $j$ to $i$ or from $i$ to $j$, respectively. If either Local-Prev-Support$(i, a)$ or Local-Next-Support$(i, a)$ becomes empty, then $a$ is no longer a part of any MUSE arc consistent instance, and should be eliminated from $L_i$. In Figure 20, the label $a$ is admissible for the segments containing both $i$ and $j$, but not for the segments containing $i$ and $k$. If because of constraints, the labels in $j$ become inconsistent with $a$ on $i$, $(i, j)$ would be eliminated from Local-Next-Support$(a, i)$, leaving an empty set. In that case, $a$ would no longer be supported by any segment.

The algorithm can utilize similar conditions for nodes which are not directly connected to $i$ by Next-Edge$_i$ or Prev-Edge$_i$. Consider Figure 21. Suppose that the label $a$ at node $i$ is compatible with a label in $L_j$, but it is incompatible with the labels in $L_x$ and $L_y$, then it is reasonable to eliminate $a$ for all segments containing both $i$ and $j$, because those segments would have to include either node $x$ or $y$. To determine whether a label is admissible for a set of segments containing $i$ and $j$, we calculate Prev-Support$[(i, j), a]$ and Next-Support$[(i, j), a]$ sets. Next-Support$[(i, j), a]$ includes all $(i, k)$ arcs which support $a$ in $i$ given that there is a directed edge from $j$ to $k$, and $(i, j)$ supports $a$. Prev-Support$[(i, j), a]$ includes all $(i, k)$ arcs which support $a$ in $i$ given that there is a directed edge from $k$ to $j$, and $(i, j)$ supports $a$. Note that Prev-Support$[(i, j), a]$ will contain an ordered pair $(i, j)$ if $(i, j) \in$ Prev-Edge$_j$, and Next-Support$[(i, j), a]$ will contain an ordered pair $(i, j)$ if $(j, i) \in$ Next-Edge$_j$. These elements are included because the edge between nodes $i$ and $j$ is sufficient to allow $j$'s labels to support $a$ in the segment containing $i$ and $j$. Dummy ordered pairs are also created to handle cases where a node is at the beginning or end of a network: when $(\text{start}, j) \in$ Prev-Edge$_j$, $(i, \text{start})$ is added to Prev-Support$[(i, j), a]$, and when $(j, \text{end}) \in$ Next-Edge$_j$, $(i, \text{end})$ is added to Next-Support$[(i, j), a]$. This is to prevent a label from being ruled out because no nodes precede or follow it in the DAG.

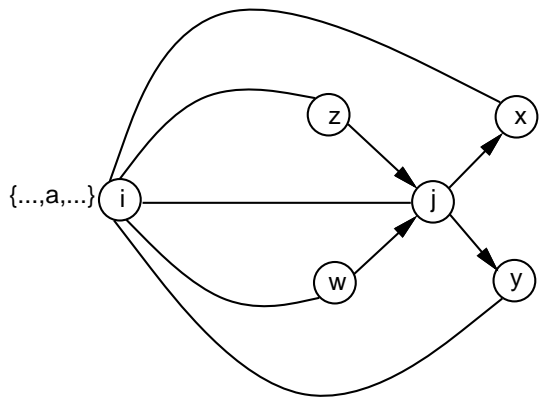

Figure 21: If Next-Edge$_j$ = $\{(j, x), (j, y)\}$, Counter$[(i, x), a] = 0$, and Counter$[(i, y), a] = 0$, then $a$ is inadmissible for every segment containing both $i$ and $j$. The solid directed lines are members of $G$, and the solid undirected lines represent members of $E$.

Figure 22 shows the Prev-Support, Next-Support, Local-Next-Support, and Local-Prev-Support sets that the initialization algorithm creates for a simple example DAG. After the initialization step, these sets contain all node pairs that are allowed based on the connectivity of $G$. Later, during the consistency step those node pairs which do not support the associated label are eliminated from each set.

To illustrate how these data structures are used by the second step of MUSE AC-1 shown in Figure 23, consider what happens if initially $[(1, 3), a] \in List$ for the MUSE CSP depicted in Figure 22. $[(1, 3), a]$ is placed on $List$ to indicate that the label $a$ in $L_1$ is not supported by any of the labels associated with node 3. When that value is popped off $List$, it is necessary for each $(3, x) \in$ S$[(1, 3), a]$ to decrement Counter$[(3, 1), x]$ by one. If any Counter$[(3, 1), x]$ becomes 0, and $[(3, 1), x]$ has not already been placed on the $List$, then it is added for future processing. Once this is done, it is necessary to remove $[(1, 3), a]$'s influence on the MUSE DAG. To handle this, we examine the two sets Prev-Support$[(1, 3), a] = \{(1, 2), (1, 3)\}$ and

1.  $List := \phi;$
2.  $E := \{(i,j) | \exists \sigma \in \Sigma : i,j \in \sigma \wedge i \neq j \wedge i,j \in N\};$
3.  **for** $(i,j) \in E$ **do**
4.      **for** $a \in L_i$ **do** {
5.          $S[(i,j),a] := \phi;$
6.          $M[(i,j),a] := 0;$
7.          Local-Prev-Support$(i,a) := \phi;$ Local-Next-Support$(i,a) := \phi;$
8.          Prev-Support$[(i,j),a] := \phi;$ Next-Support$[(i,j),a] := \phi;$ }
9.  **for** $(i,j) \in E$ **do**
10.     **for** $a \in L_i$ **do** {
11.         Total $:= 0;$
12.         **for** $b \in L_j$ **do**
13.             **if** $R2(i,a,j,b)$ **then** {
14.                Total $:=$ Total$+1;$
15.                $S[(j,i),b] := S[(j,i),b] \cup \{(i,a)\};$ }
16.         **if** Total$=0$ **then** {
17.          $List := List \cup \{[(i,j),a]\};$
18.         $M[(i,j),a] := 1;$ }
19.         Counter$[(i,j),a] :=$ Total;
20.         Prev-Support$[(i,j),a] := \{(i,x) | (i,x) \in E \wedge (x,j) \in \text{Prev-Edge}_j\}$
                                        $\cup \{(i,j) | (i,j) \in \text{Prev-Edge}_j\}$
                                        $\cup \{(i,\text{start}) | (\text{start},j) \in \text{Prev-Edge}_j\};$
21.         Next-Support$[(i,j),a] := \{(i,x) | (i,x) \in E \wedge (j,x) \in \text{Next-Edge}_j\}$
                                        $\cup \{(i,j) | (j,i) \in \text{Next-Edge}_j\}$
                                        $\cup \{(i,\text{end}) | (j,\text{end}) \in \text{Next-Edge}_j\};$
22.         Local-Prev-Support$(i,a) := \{(i,x) | (i,x) \in E \wedge (x,i) \in \text{Prev-Edge}_i\}$
                                        $\cup \{(i,\text{start}) | (\text{start},i) \in \text{Prev-Edge}_i\};$
23.         Local-Next-Support$(i,a) := \{(i,x) | (i,x) \in E \wedge (i,x) \in \text{Next-Edge}_i\}$
                                        $\cup \{(i,\text{end}) | (i,\text{end}) \in \text{Next-Edge}_i\};$ }

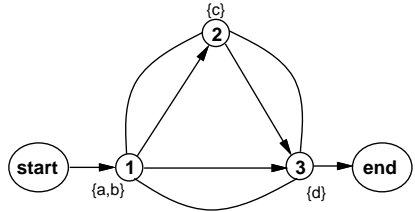

| | |
|---|---|
| Prev-Support$[(1,2),a] = \{(1,2)\}$ | Next-Support$[(1,2),a] = \{(1,3)\}$ |
| Prev-Support$[(1,3),a] = \{(1,2),(1,3)\}$ | Next-Support$[(1,3),a] = \{(1,\text{end})\}$ |
| Prev-Support$[(1,2),b] = \{(1,2)\}$ | Next-Support$[(1,2),b] = \{(1,3)\}$ |
| Prev-Support$[(1,3),b] = \{(1,2),(1,3)\}$ | Next-Support$[(1,3),b] = \{(1,\text{end})\}$ |
| Prev-Support$[(2,1),c] = \{(2,\text{start})\}$ | Next-Support$[(2,1),c] = \{(2,1),(2,3)\}$ |
| Prev-Support$[(2,3),c] = \{(2,3),(2,1)\}$ | Next-Support$[(2,3),c] = \{(2,\text{end})\}$ |
| Prev-Support$[(3,1),d] = \{(3,\text{start})\}$ | Next-Support$[(3,1),d] = \{(3,1),(3,2)\}$ |
| Prev-Support$[(3,2),d] = \{(3,1)\}$ | Next-Support$[(3,2),d] = \{(3,2)\}$ |
| Local-Prev-Support$(1,a) = \{(1,\text{start})\}$ | Local-Next-Support$(1,a) = \{(1,2),(1,3)\}$ |
| Local-Prev-Support$(1,b) = \{(1,\text{start})\}$ | Local-Next-Support$(1,b) = \{(1,2),(1,3)\}$ |
| Local-Prev-Support$(2,c) = \{(2,1)\}$ | Local-Next-Support$(2,c) = \{(2,3)\}$ |
| Local-Prev-Support$(3,d) = \{(3,1),(3,2)\}$ | Local-Next-Support$(3,d) = \{(3,\text{end})\}$ |

Figure 22: Initialization of the data structures for MUSE AC-1 along with a simple example.

```
1.    while List ≠ φ do {
2.         Pop [[(i, j), a] from List;
3.         for (j, b) ∈ S[(i, j), a] do {
4.              Counter[(j, i), b] := Counter[(j, i), b] − 1;
5.              if Counter[(j, i), b] = 0 ∧ M[(j, i), b] = 0 then {
6.                   List := List ∪ {[(j, i), b]};
7.                   M[(j, i), b] := 1; } }
8.         Update-Support-Sets([(i, j), a]); (see Figure 24) }
```

Figure 23: Eliminating inconsistent labels from the domains in MUSE AC-1.

```
Update-Support-Sets ([(i, j), a]) {
1.    for (i, x) ∈ Prev-Support[(i, j), a] ∧ x ≠ j ∧ x ≠ start do {
2.         Prev-Support[(i, j), a] := Prev-Support[(i, j), a] − {(i, x)} ;
3.         Next-Support[(i, x), a] := Next-Support[(i, x), a] − {(i, j)};
4.         if Next-Support[(i, x), a] = φ ∧ M[(i, x), a] = 0 then {
5.              List := List ∪ {[(i, x), a]};
6.              M[(i, x), a] := 1; } }
7.    for (i, x) ∈ Next-Support[(i, j), a] ∧ x ≠ j ∧ x ≠ end do {
8.         Next-Support[(i, j), a] := Next-Support[(i, j), a] − {(i, x)};
9.         Prev-Support[(i, x), a] := Prev-Support[(i, x), a] − {(i, j)};
10.        if Prev-Support[(i, x), a] = φ ∧ M[(i, x), a] = 0 then {
11.             List := List ∪ {[(i, x), a]};
12.             M[(i, x), a] := 1; } }
13.   if (j, i) ∈ Prev-Edge_i then
14.        Local-Prev-Support(i, a) := Local-Prev-Support(i, a) − {(i, j)};
15.   if Local-Prev-Support(i, a) = φ then {
16.        L_i := L_i − {a};
17.        for (i, x) ∈ Local-Next-Support(i, a) ∧ x ≠ j ∧ x ≠ end do {
18.             Local-Next-Support(i, a) := Local-Next-Support(i, a) − {(i, x)};
19.             if M[(i, x), a] = 0 then {
20.                  List := List ∪ {[(i, x), a]};
21.                  M[(i, x), a] := 1; } } }
22.   if (i, j) ∈ Next-Edge_i then
23.        Local-Next-Support(i, a) := Local-Next-Support(i, a) − {(i, j)};
24.   if Local-Next-Support(i, a) = φ then {
25.        L_i := L_i − {a};
26.        for (i, x) ∈ Local-Prev-Support(i, a) ∧ x ≠ j ∧ x ≠ start do {
27.             Local-Prev-Support(i, a) := Local-Prev-Support(i, a) − {(i, x)};
28.             if M[(i, x), a] = 0 then {
29.                  List := List ∪ {[(i, x), a]};
30.                  M[(i, x), a] := 1; } } } }
```

Figure 24: The function Update-Support-Sets([(i, j), a]) for MUSE AC-1.

Next-Support$[(1,3),a] = \{(1,\text{end})\}$. Note that the value $(1,\text{end})$ in Next-Support$[(1,3),a]$ and the value $(1,3)$ in Prev-Support$[(1,3),a]$, require no further action because they are dummy values. However, the value $(1,2)$ in Prev-Support$[(1,3),a]$ indicates that $(1,3)$ is a member of Next-Support$[(1,2),a]$, and since $a$ is not admissible for $(1,3)$, $(1,3)$ should be removed from Next-Support$[(1,2),a]$, leaving an empty set. Note that because Next-Support$[(1,2),a]$ is empty, and assuming that M$[(1,2),a] = 0$, $[(1,2),a]$ is added to *List* for further processing. Next, $(1,3)$ is removed from Local-Next-Support$(1,a)$, leaving a set of $\{(1,2)\}$. During the next iteration of the while loop $[(1,2),a]$ is popped from *List*. When Prev-Support$[(1,2),a]$ and Next-Support$[(1,2),a]$ are processed, Next-Support$[(1,2),a] = \phi$ and Prev-Support$[(1,2),a]$ contains only a dummy, requiring no action. Finally, when $(1,2)$ is removed from Local-Next-Support$(1,a)$, the set becomes empty, so $a$ is no longer compatible with any segment containing node 1 and can be eliminated from further consideration as a possible label for node 1. Once $a$ is eliminated from node 1, it is also necessary to remove the support of $a \in L_1$ from all labels on nodes that precede node 1, that is for all nodes $x$ such that $(1,x) \in$ Local-Prev-Support$(1,a)$. Since Local-Prev-Support$(1,a) = \{(1,\text{start})\}$, and start is a dummy node, there is no more work to be done.

In contrast, consider what happens if initially $[(1,2),a] \in$ *List* for the MUSE CSP in Figure 22. In this case, Prev-Support$[(1,2),a]$ contains $(1,2)$ which requires no additional work; whereas, Next-Support$[(1,2),a]$ contains $(1,3)$, indicating that $(1,2)$ must be removed from Prev-Support$[(1,3),a]$'s set. After the removal, Prev-Support$[(1,3),a]$ is non-empty, so the segment containing nodes 1 and 3 still supports the label $a$ in $L_1$. The reason that these two cases provide different results is that the constraint arc between nodes 1 and 3 is contained in every segment; whereas, the constraint arc between nodes 1 and 2 is found in only one of them.

## 3.3 The Running Time and Space Complexity of MUSE AC-1

The worst-case running time of the routine to initialize the MUSE AC-1 data structures (in Figure 22) is O$(n^2l^2 + n^3l)$, where $n$ is the number of nodes in a MUSE CSP and $l$ is the number of labels. Given that the number of $(i,j)$ elements in $E$ is O$(n^2)$ and the domain size is O$(l)$, the size of the Counter and S arrays is O$(n^2l)$. To determine the number of supporters for a given arc-label pair requires O$(l)$ work; hence, initializing the Counter and S arrays requires O$(n^2l^2)$ time. However, there are O$(n^2l)$ Prev-Support and Next-Support sets, where each Prev-Support$[(i,j),a]$ and Next-Support$[(i,j),a]$ requires O$(n)$ time to compute, so the time to calculate all Prev-Support and Next-Support sets is O$(n^3l)$. Finally, the time needed to calculate all Local-Next-Support and Local-Prev-Support sets is O$(n^2l)$ because there are O$(nl)$ sets with up to O$(n)$ elements per set.

The worst-case running time for the algorithm which prunes labels that are not MUSE arc consistent (in Figures 23 and 24) also operates in O$(n^2l^2 + n^3l)$ time. Clearly the Counter array contains O$(n^2l)$ entries (a similar argument can be made for the S array) to keep track of in the algorithm. Each Counter$[(i,j),a]$ can be at most $l$ in magnitude, and it can never become negative, so the maximum running time for line 4 in Figure 23 (given that elements appear on *List* only once because of M) is O$(n^2l^2)$. Because there are O$(n^2l)$ Next-Support and Prev-Support lists, each up to O$(n)$ in size, the maximum running time required for lines 3 and 9 in Figure 24 is O$(n^3l)$. Finally, since there are O$(nl)$

| Approach | Nodes per Path | Degree of Node splitting | Number of Constraint Networks | Number of Nodes | Asymptotic Time |
|---|---|---|---|---|---|
| CSPs | $n$ | $k$ | $k^n$ | $n$ | $k^n n^2 l^2$ |
| MUSE CSP | $n$ | $k$ | 1 | $kn$ | $(kn)^2 l^2 + (kn)^3 l$ |

Table 1: Comparison of the space and time complexity for MUSE arc consistency on a MUSE CSP to arc consistency on multiple CSPs representing a node splitting problem (e.g., lexical ambiguity in parsing).

Local-Prev-Support and Local-Next-Support sets from which to eliminate $O(n)$ elements, the maximum running time of lines 14 and 23 in Figure 24 is $O(n^2 l)$. Hence, the maximum running time of the MUSE CSP arc consistency algorithm is $O(n^2 l^2 + n^3 l)$.

The space complexity of MUSE CSP AC-1 is also $O(n^2 l^2 + n^3 l)$ because the arrays Counter and M contain $O(n^2 l)$ elements, and there are $O(n^2 l)$ S sets, each containing $O(l)$ items; $O(n^2 l)$ Prev-Support and Next-Support sets, each containing $O(n)$ items; and $O(nl)$ Local-Next-Support and Local-Prev-Support sets, each containing $O(n)$ items.

By comparison, the worst-case running time and space complexity for CSP arc consistency is $O(n^2 l^2)$, assuming that there are $n^2$ constraint arcs. Note that for applications where $l = n$, the worst-case running times of the algorithms are the same order (this is true for parsing spoken language with a MUSE CSP). Also, if $\Sigma$ is representable as planar DAG (in terms of Prev-Edge and Next-Edge, not E), then the running times of the two algorithms are the same order because the average number of values in Prev-Support and Next-Support would be a constant. On the other hand, if we compare MUSE CSP to the use of multiple CSPs for problems where there are $k$ alternative variables for a particular variable in a CSP, then MUSE CSP AC-1 is asymptotically more attractive, as shown in Table 1.

## 3.4 The Correctness of MUSE AC-1

Next we prove the correctness of MUSE AC-1.

**Theorem 1** *A label $a$ is eliminated from $L_i$ by MUSE AC-1 if and only if that label is unsupported by all the arcs $(i, x)$ of every segment.*

**Proof:**

1. We must show that if a label is eliminated, it is inadmissible in every segment. A label is eliminated from a domain by MUSE AC-1 (see lines 16 and 25 in Figure 24) if and only if its Local-Prev-Support set or its Local-Next-Support set becomes empty (see lines 15 and 24 in Figure 24). In either case, the label should be eliminated to make the MUSE CSP instance MUSE arc consistent. We prove that if a label's local support sets become empty, that label cannot participate in any MUSE arc consistent instance of MUSE CSP. This is proven for Local-Next-Support (Local-Prev-Support follows by symmetry.) Observe that if $a \in L_i$, and it is unsupported by

266

all of the nodes which immediately follow $i$ in the DAG, then it cannot participate in any MUSE arc consistent instance of MUSE CSP. In line 23 of Figure 24, if $(i, j)$ is removed from Local-Next-Support$(i, a)$ set then $[(i, j), a]$ must have been popped off $List$. The removal of $(i, j)$ from Local-Next-Support$(i, a)$ indicates that, in the segment containing $i$ and $j$, $a \in L_i$ is inadmissible. It remains to be shown that $[(i, j), a]$ is put on $List$ if $a \in L_i$ is unsupported by every segment which contains $i$ and $j$. This is proven by induction on the number of iterations of the while loop in Figure 23.

**Base case:** The initialization routine only puts $[(i, j), a]$ on $List$ if $a \in L_i$ is incompatible with every label in $L_j$ (line 17 of Figure 22). Therefore, $a \in L_i$ is unsupported by all segments containing $i$ and $j$.

**Induction step:** Assume that at the start of the $k$th iteration of the while loop all $[(x, y), c]$ which have ever been put on $List$ indicate that $c \in L_x$ is inadmissible in every segment which contains $x$ and $y$. It remains to show that during the $k$th iteration, if $[(i, j), a]$ is put on $List$, then $a \in L_i$ is unsupported by every segment which contains $i$ and $j$. There are several ways in which a new $[(i, j), a]$ can be put on $List$:

(a) All labels in $L_j$ which were once compatible with $a \in L_i$ have been eliminated. This item could have been placed on $List$ either during initialization (see line 17 in Figure 22) or during a previous iteration of the while loop (see line 6 in Figure 23)), just as in the CSP AC-4 algorithm. It is obvious that, in this case, $a \in L_i$ is inadmissible in every segment containing $i$ and $j$.

(b) Prev-Support$[(i, j), a] = \phi$ (see line 10 in Figure 24) indicating that $a \in L_i$ is incompatible with all nodes $k$ for $(k, j) \in$ Prev-Edge$_j$. The only way for $[(i, j), a]$ to be placed on $List$ for this reason (at line 11) is because all tuples of the form $[(i, k), a]$ (where $(k, j) \in$ Prev-Edge$_j$) were already put on $List$. By the induction hypothesis, these $[(i, k), a]$ items were placed on the $List$ because $a \in L_i$ is inadmissible in with all segments containing $i$ and $k$ in the DAG. But if $a$ is not supported by any node which immediately precedes $j$ in the DAG, then $a$ is unsupported by every segment which contains $j$. Therefore, it is correct to put $[(i, j), a]$ on $List$.

(c) Next-Support$[(i, j), a] = \phi$ (see line 4 in Figure 24) indicating that $a \in L_i$ is incompatible with all nodes $k$ for $(j, k) \in$ Next-Edge$_j$. The only way for $[(i, j), a]$ to be placed on $List$ (at line 5) for this reason is because all tuples of the form $[(i, k), a]$ (where $(j, k) \in$ Next-Edge$_j$) were already put on $List$. By the induction hypothesis, these $[(i, k), a]$ items were placed on the $List$ because $a \in L_i$ is inadmissible in all segments containing $i$ and $k$ in the DAG. But if $a$ is not supported by any node which immediately follows $j$ in the DAG, then $a$ is inadmissible in every segment which contains $j$. Therefore, it is correct to put $[(i, j), a]$ on $List$.

(d) Local-Next-Support$(i, a) = \phi$ (see line 24 in Figure 24) indicating that $a \in L_i$ is incompatible with all nodes $k$ such that $(i, k) \in$ Next-Edge$_i$. The only way for $[(i, j), a]$ to be placed on $List$ (at line 29) for this reason is because no node which follows $i$ in the DAG supports $a$, and so all pairs $(i, k)$ have been legally removed from Local-Next-Support$(i, a)$ during previous iterations. Because there is no segment containing $i$ which supports $a$, it follows that no segment containing $i$ and $j$ supports that label.

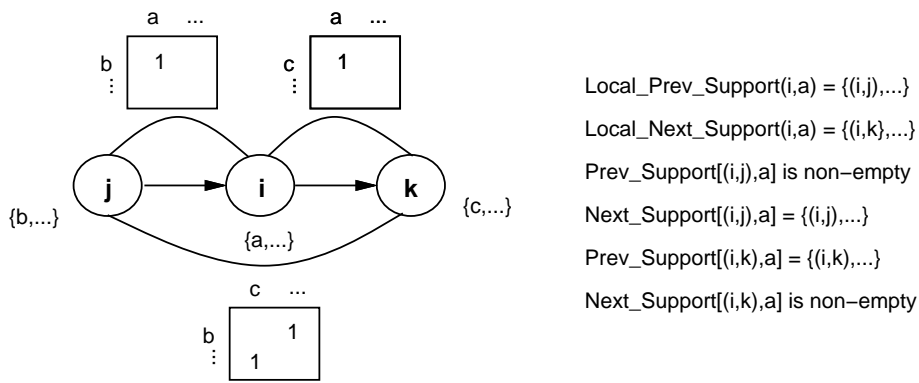

Figure 25: If $a \in L_i$ after MUSE AC-1, it must be preceded by some node $j$ and followed by some node $k$ which support $a$.

(e) Local-Prev-Support$(i, a) = \phi$ (see line 15 in Figure 24) indicating that $a \in L_i$ is incompatible with all nodes $k$ such that $(k, i) \in$ Prev-Edge$_i$. The only way for $[(i, j), a]$ to be placed on $List$ (at line 20) for this reason is because no node which precedes $i$ in the DAG supports $a$, and so all pairs $(i, k)$ have been legally removed from Local-Prev-Support$(i, a)$ during previous iterations. Because there is no segment containing $i$ which supports $a$, it follows that no segment containing $i$ and $j$ supports that label.

At the beginning of the $(k + 1)$th iteration of the while loop, every $[(x, y), c]$ on $List$ implies that $c$ is not supported by any segment which contains $x$ and $y$. Therefore, by induction, it is true for all iterations of the while loop in Figure 23. Hence, if a label's local support sets become empty, that label cannot participate in a MUSE arc consistent instance of MUSE CSP.

2. We must also show that if $a$ is not eliminated from $L_i$ by the MUSE arc consistency algorithm, then it must be MUSE arc consistent. For $a$ to be MUSE arc consistent, there must exist at least one path from start to end which goes through node $i$ such that all nodes $n$ on that path contain at least one label which is compatible with $a \in L_i$. If $a$ is not deleted after MUSE AC-1, then Local-Next-Support$(i, a) \neq \phi$ and Local-Prev-Support$(i, a) \neq \phi$. Hence, $i$ must be preceded and followed by at least one node which supports $a \in L_i$; otherwise, $a$ would have been deleted. As depicted in Figure 25, we know that there must be some node $j$ which precedes $i$ such that, if it is not start, it must contain at least one label $b$ which supports $a$, and Next-Support$[(i, j), a]$ and Prev-Support$[(i, j), a]$ must be non-empty. Similarly, there must be some node $k$ which follows $i$ such that, if it is not end, it must contain at least one label $c$ which supports $a$, and Next-Support$[(i, k), a]$ and Prev-Support$[(i, k), a]$ must be non-empty.

To show there is a path through the DAG, we must show that there is a path beginning at start which reaches $i$ such that all the nodes along that path support $a \in L_i$, and that there is a path beginning at $i$ which reaches end such that all the nodes along that path support $a \in L_i$. We will show the necessity of the path from $i$ to end such that all nodes along that path support $a \in L_i$ given that $a$ remains after MUSE AC-1; the necessity of the path from start to $i$ can be shown in a similar way.

**Base case:** If $a \in L_i$ after MUSE AC-1, then there must exist at least one node which follows $i$, say $k$, such that $[(i,k),a]$ has never been placed on *List*. Hence, $R2(i,a,k,c) = 1$ for at least one $c \in L_k$ and Next-Support$[(i,k),a]$ and Prev-Support$[(i,k),a]$ must be non-empty.

**Induction Step:** Assume that there is a path of $n$ nodes that follows $i$ that supports $a \in L_i$, but none of those nodes is the end node. This implies that each of the $n$ nodes contains at least one label compatible with $a$ and that Next-Support$[(i,n),a]$ and Prev-Support$[(i,n),a]$ must be non-empty for each of the $n$ nodes.

Next, we show that a path of length $(n + 1)$ must also support $a \in L_i$; otherwise, the label $a$ would have been deleted by MUSE AC-1. We have already noted that for the $n$th node on the path in the induction step, Next-Support$[(i,n),a]$ must be non-empty; hence, there must exist at least one node, say $n'$, which follows the $n$th node in the path of length $n$ which supports $a \in L_i$. If $n'$ is the end node, then this is the case. If $n'$ is not end, then the only way that $(i,n')$ can be a member of Next-Support$[(i,n),a]$ is if $[(i,n'),a]$ has not been placed on *List*. If it hasn't, then $R2(i,a,n',l) = 1$ for at least one $l \in L_{n'}$ and Next-Support$[(i,n'),a]$ and Prev-Support$[(i,n'),a]$ must be non-empty. If this were not the case, then $(i,n')$ would have been removed from Next-Support$[(i,n),a]$, and $n$ would no longer support $a \in L_i$.

Hence, if $a \in L_i$ after MUSE AC-1, then there must be a path of nodes to end such that for each node $n$ which is not the end node, $R2(i,a,n,l) = 1$ for at least one $l \in L_n$ and Next-Support$[(i,n),a]$ and Prev-Support$[(i,n),a]$ must be non-empty. Hence $a$ is MUSE arc consistent.
□

From this theorem, we may conclude that MUSE AC-1 builds the largest MUSE arc consistent structure. Because MUSE arc consistency takes into account all of its segments, if a single CSP were selected from the MUSE CSP after MUSE arc consistency is enforced, CSP arc consistency could eliminate additional labels.

## 3.5 A Profile of MUSE AC-1

Given the fact that MUSE AC-1 operates on a composite data structure, the benefits of using this algorithm can have a high payoff over individually processing CSPs. In section 2.4, we provided several examples where the payoff is obvious. To gain some insight into factors influencing the effectiveness of MUSE CSP, we have conducted an experiment in which we randomly generate MUSE CSP instances with two different graph topologies. The *tree* topology is characterized by two parameters: the *branching factor* (how many nodes follow each non-leaf node in the tree) and the *path length* (how many nodes there are in a path from the root node to a leaf node). The *lattice* topology is characteristic of a MUSE CSP which is produced by a hidden-Markov-model-based spoken language recognition system for our constraint-based parser. Lattices are also characterized by their length and their branching factor.

For this experiment, we examined trees with a path length of four and a branching factor of two or three, and lattices with a path length of four and a branching factor of two or three. We initialized each variable to have either 3 or 6 labels. We then randomly generated constraints in the network, varying the probability that $R2(i, a, j, b) = 1$ from 0.05 to .95 in steps of 0.05. For each probability, 6 instances were generated. The lower the probability that $R2(i, a, j, b) = 1$, the tighter the constraints. Note that the probability of a constraint between two nodes should be understood as the probability of a constraint between two nodes given that a constraint is allowed between them. For example, nodes that are on the same level in the tree topology are in different segments, and so constraints cannot occur between them.

The results of this experiment are displayed in Figures 26 and 27. In each of the four panels of each figure, four curves are displayed. *After MUSE AC-1* appears on curves displaying the average number of labels remaining after MUSE AC-1 is applied to instances of a MUSE CSP as the probability of a constraint varies. The curves labeled *Solution* indicate the average number of labels remaining after MUSE AC-1 that are used in a solution. *CSP AC* is associated with curves that display the number of labels that remain in at least one segment when the segment is extracted from the MUSE CSP and CSP arc consistency is applied. *Unused* indicates the difference between the number of labels that remain after MUSE AC-1 and the number that are CSP arc consistent in at least one segment.

For both of the topologies, if the probability $R2(i, a, j, b) = 1$ is low (e.g., .1) or high (e.g., .8), then MUSE AC-1 tracks the performance of arc consistency performed on the individual instances for either topology. However, the topology does impact the range of low and high probabilities for which this is true. When constraints are randomly generated, after MUSE AC-1 is performed, the tree topology has fewer remaining values than lattice topology that are not CSP arc consistent. These results suggest that MUSE CSP AC-1 may be more effective for some topologies than for others. However, in the tree topology the randomly generated constraints between the values of two variables are independent of the other probabilities generated. This is not the case for the lattice; once a pair of variables has a set of randomly generated constraints, they are shared by all paths through the lattice. Notice that increasing the number of values in a domain seems to have more impact on the tree than increasing the branching factor, probably because as the branching factor increases, so does the number of independent nodes.

This experiment does show that if a problem is tightly constrained, MUSE AC-1 can be effectively used to eliminate values that are unsupported by the constraints. Clearly, this was the case for the parsing problems presented in section 2.4. A small set of syntactic constraints effectively eliminates values that can never be used in a parse for a sentence, even in a lattice with a branching factor of three and arbitrarily long paths.

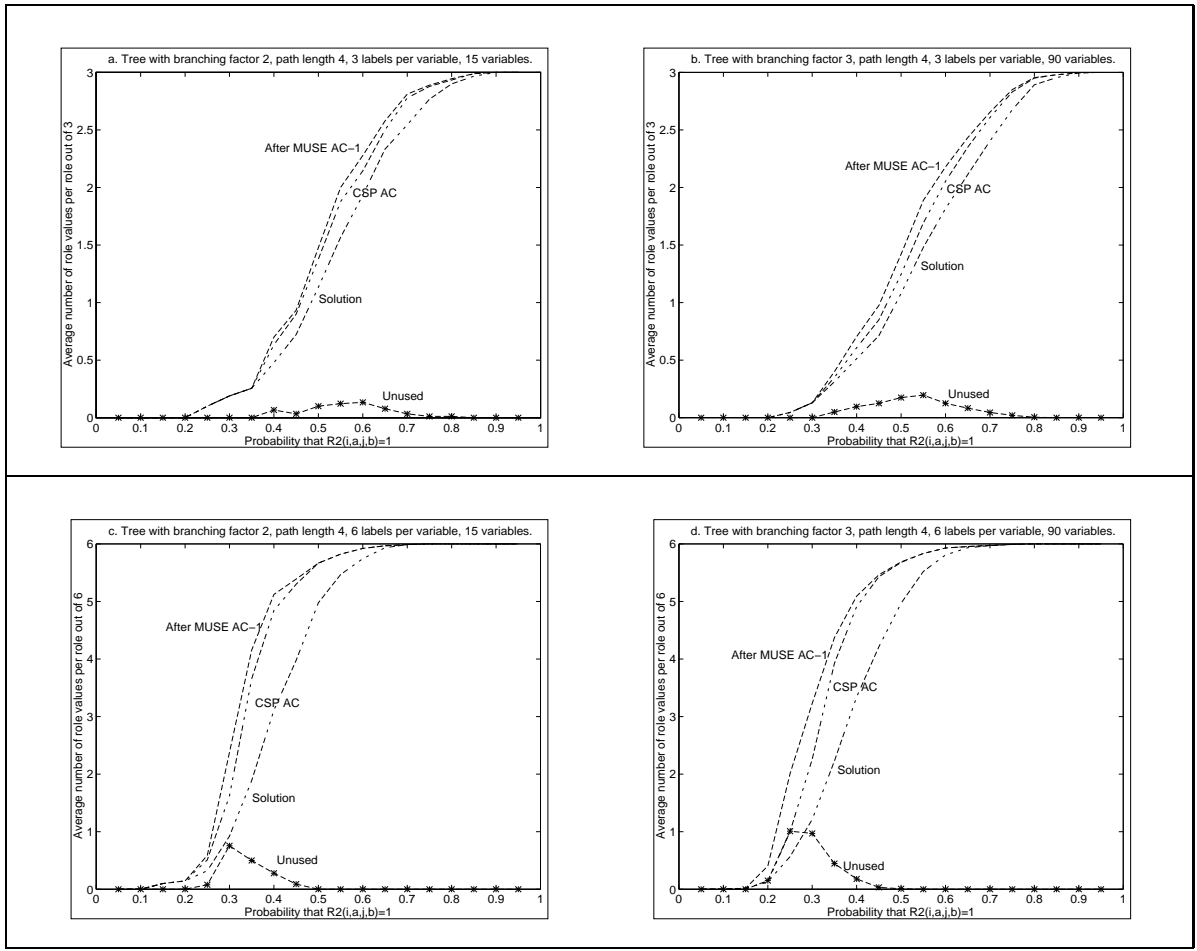

Figure 26: Simulation results for trees with a path length of 4, a branching factor of 2 or 3, and 3 or 6 labels per variable.

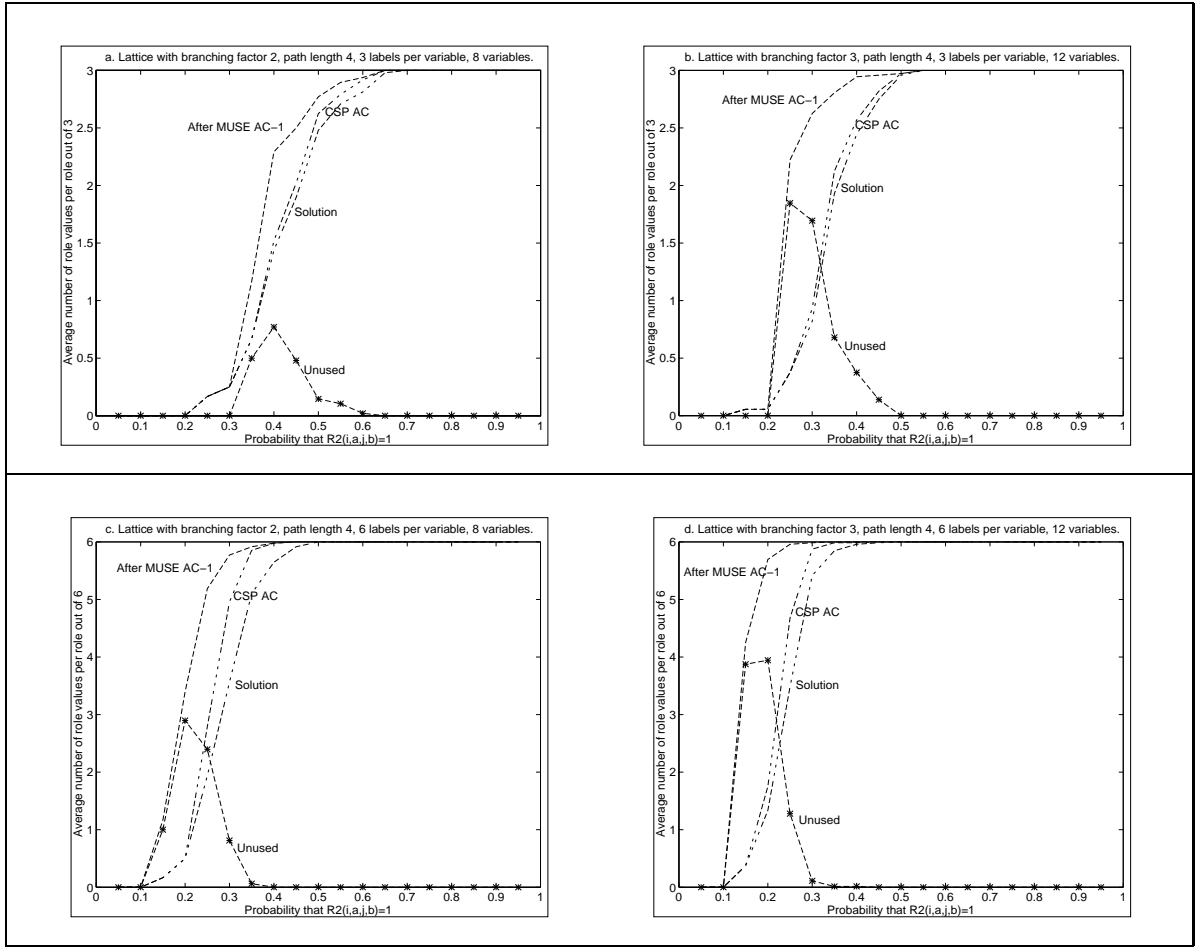

Figure 27: Simulation results for lattices with a path length of 4, a branching factor of 2 or 3, and 3 or 6 labels per variable.

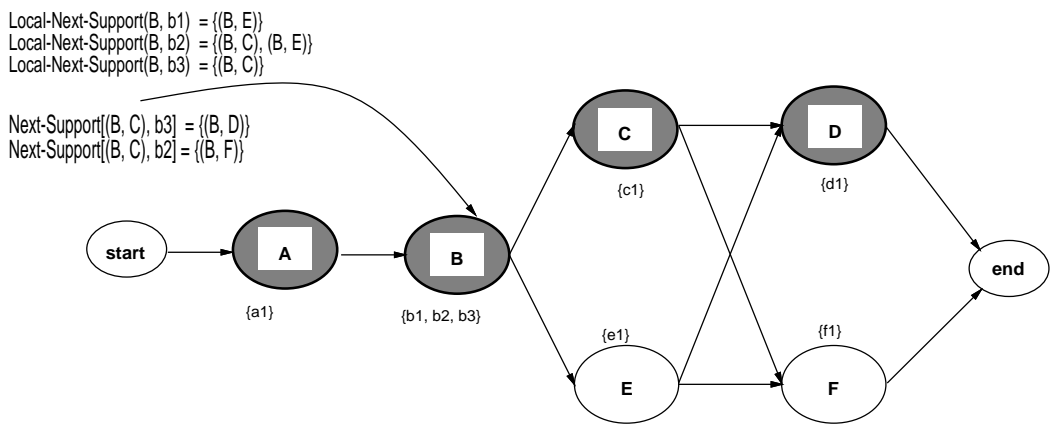

Local-Next-Support(B, b1) = {(B, E)}
Local-Next-Support(B, b2) = {(B, C), (B, E)}
Local-Next-Support(B, b3) = {(B, C)}

Next-Support[(B, C), b3] = {(B, D)}
Next-Support[(B, C), b2] = {(B, F)}

Figure 28: Using MUSE arc consistency data structures to guide a backtracking search.

## 3.6 Extracting Solutions from a MUSE CSP after MUSE AC-1

Solutions to regular CSP problems are typically generated by using backtracking (or fancier search algorithms) to assemble a set of labels, one for each node, which are consistently admissible. Extracting solutions from MUSE CSPs can be done in a similar way, but it is desirable to make a few modifications to the search algorithms to take advantage of the extra information which is contained in the MUSE AC-1 data structures.

Consider the example shown in Figure 28. This figure presents a simple MUSE CSP. Suppose we are only interested in solutions to the segment which is highlighted: {A, B, C, D}. Suppose also that there is only one solution to this segment: a1 for A, b3 for B, c1 for C, and d1 for D. We wish to find this solution by depth-first search.

We begin by assigning a1 to A. However, the domain of B, in addition to the desired label b3, also contains the labels b1 and b2, which are valid only for other segments. If we initially (and naively) choose b1 for B and continue doing depth-first search, we would waste a lot of time backtracking. Fortunately, after enforcing MUSE arc consistency, the MUSE data structures contain useful information concerning the segments for which the labels are valid. In this case, the backtracking algorithm can check Local-Next-Support(B, b1) to determine which of the outgoing nodes b1 is compatible with. Since (B, C) is not an element of Local-Next-Support(B, b1), a smart search algorithm would not choose b1 as a label for B.

However, just looking at the local support sets might not be enough. After the search algorithm has rejected b1 as a label for B, it would go on to consider b2. Local-Next-Support(B, b2) indicates that b2 is a valid label for some of the segments which contain C, but it fails to tell us that b2 is not valid for the segment we are examining. Despite this, the search algorithm can still eliminate b2 by looking at Next-Support[(B, C), b2], which indicates that b2 is only compatible with segments containing the node F. Clearly, this type of information will more effectively guide the search for a solution along a certain path. Improved search strategies for MUSE CSPs will be the focus of future research efforts.

## 4. The MUSE CSP Path Consistency Algorithm

In this section, we introduce an algorithm to achieve MUSE CSP path consistency, MUSE PC-1, which builds upon the PC-4 algorithm (Han & Lee, 1988).

### 4.1 MUSE PC-1

MUSE path consistency is enforced by setting $R2(i, a, j, b)$ to false when it violates the conditions of Definition 9. MUSE PC-1 builds and maintains several data structures comparable to the data structures defined for MUSE AC-1, described in Figure 29, to allow it to efficiently perform this operation. Figure 32 shows the code for initializing the data structures, and Figures 33 and 34 contain the algorithm for eliminating MUSE path inconsistent binary constraints.

MUSE PC-1 must keep track of which labels in $L_k$ support $R2(i, a, j, b)$. To keep track of how much path support each $R2(i, a, j, b)$ has, the number of labels in $L_k$ which satisfy $R2(i, a, k, c)$ and $R2(k, c, j, b)$ are counted using Counter$[(i, j), k, a, b]$. Additionally, the algorithm must keep track of the set S$[(i, j), k, a, b]$, which contains members of the form $(k, c)$ where $R2(i, a, k, c)$ and $R2(k, c, j, b)$ are supported by $R2(i, a, j, b)$. If $R2(i, a, j, b)$ ever becomes false in the segment containing $i$, $j$, and $k$, then $R2(i, a, k, c)$ and $R2(k, c, j, b)$ will loose some of their support. MUSE PC-1 also uses the Local-Next-Support, Local-Prev-Support, Prev-Support, and Next-Support sets similar to those in MUSE AC-1.

MUSE PC-1 is able to use the properties of the DAG to identify local (and hence efficiently computable) conditions under which binary constraints fail because of lack of path support. Consider Figure 30, which shows the nodes which are adjacent to node $i$ and $j$ in the DAG. Because every segment in the DAG which contains node $i$ and $j$ is represented as a directed path in the DAG going through both node $i$ and node $j$, some node must precede and follow nodes $i$ and $j$ for $R2(i, a, j, b)$ to hold. In order to track this dependency, two sets are maintained for each $[(i, j), a, b]$ tuple: Local-Prev-Support$[(i, j), a, b]$ and Local-Next-Support$[(i, j), a, b]$. Note that we distinguish Local-Prev-Support$[(i, j), a, b]$ from Local-Prev-Support$[(j, i), b, a]$ to separately keep track of those elements directly preceding $i$ and those directly preceding $j$. We also distinguish Local-Next-Support$[(i, j), a, b]$ from Local-Next-Support$[(j, i), b, a]$. If any of these sets become empty, then the $(i, j)$ arc can no longer support $R2(i, a, j, b)$. Local-Prev-Support$[(i, j), a, b]$ is a set of ordered node pairs $(i, x)$ such that $(x, i) \in$ Prev-Edge$_i$, and if $(i, x) \in E$, there is at least one label $d \in L_x$ which is compatible with $R2(i, a, j, b)$. Local-Next-Support$[(i, j), a, b]$ is a set of ordered node pairs $(i, x)$ such that $(i, x) \in$ Next-Edge$_i$, and if $(i, x) \in E$, there is at least one label $d \in L_x$ which is compatible with $R2(i, a, j, b)$. Dummy ordered pairs are also created to handle cases where a node is at the beginning or end of a network: when $(start, i) \in$ Prev-Edge$_i$, $(i, start)$ is added to Local-Prev-Support$[(i, j), a, b]$, and when $(i, end) \in$ Next-Edge$_i$, $(i, end)$ is added to Local-Next-support$[(i, j), a, b]$.

The algorithm can utilize similar conditions for nodes which may not be directly connected to $i$ and $j$. Consider Figure 31. Suppose that $R2(i, a, j, b)$ is compatible with a label in $L_k$, but is incompatible with the labels in $L_x$ and $L_y$, then $R2(i, a, j, b)$ and $R2(j, b, i, a)$ are false for all segments containing $i$, $j$, and $k$ because those segments would have to include either node $x$ or $y$. To determine whether a constraint is admissible for a set of segments containing $i$, $j$, and $k$, we calculate Prev-Support$[(i, j), k, a, b]$, Prev-

| Notation | Meaning |
|---|---|
| $(i, j)$ | An ordered pair of nodes. |
| $E$ | All node pairs $(i, j)$ such that there exists a path of directed edges in $G$ between $i$ and $j$. If $(i, j) \in E$, then $(j, i) \in E$. |
| $[(i, j), k, a, b]$ | An ordered quadruple of a node pair $(i, j)$, a node $k$, and the labels $a \in L_i$ and $b \in L_j$. |
| $L_i$ | $\{a \mid a \in L$ and $(i, a)$ is permitted by the constraints (i.e., admissible)$\}$ |
| $R2(i, a, j, b)$ | $R2(i, a, j, b) = 1$ indicates the admissibility $a \in L_i$ and $b \in L_j$ given binary constraints. |
| Counter$[(i, j), k, a, b]$ | The number of labels in $L_k$ which are compatible with $R2(i, a, j, b)$. |
| S$[(i, j), k, a, b]$ | $(k, c) \in S[(i, j), k, a, b]$ means that $c \in L_k$ is compatible with $R2(i, a, j, b)$. |
| M$[(i, j), k, a, b]$ | M$[(i, j), k, a, b] = 1$ indicates that $R2(i, a, j, b)$ is false for paths including $i$, $j$, and $k$. |
| *List* | A queue of path support to be deleted. |
| $G$ | $G$ is the set of node pairs $(i, j)$ such that there exists a directed edge from $i$ to $j$. |
| Next-Edge$_i$ | Next-Edge$_i$ contains all node pairs $(i, j)$ such that there exists a directed edge $(i, j) \in G$. It also contains $(i, \text{end})$ if $i$ is the last node in a segment. |
| Prev-Edge$_i$ | Prev-Edge$_i$ contains all node pairs $(j, i)$ such that there exists a directed edge $(j, i) \in G$. It also contains $(\text{start}, i)$ if $i$ is the first node in a segment. |
| Local-Prev-Support$[(i, j), a, b]$ | A set of elements $(i, k)$ such that $(k, i) \in$ Prev-Edge$_i$, and if $k \neq \text{start}$, $R2(i, a, j, b)$ must be compatible with one of $k$'s labels. If Local-Prev-Support$[(i, j), a, b]$ becomes empty, $R2(i, a, j, b)$ becomes false. |
| Local-Next-Support$[(i, j), a, b]$ | A set of elements $(i, k)$ such that $(i, k) \in$ Next-Edge$_i$, and if $k \neq \text{end}$, $R2(i, a, j, b)$ must be compatible with one of $k$'s labels. If Local-Next-Support$[(i, j), a, b]$ becomes empty, $R2(i, a, j, b)$ becomes false. |
| Prev-Support$[(i, j), k, a, b]$ | $(i, x) \in$ Prev-Support$[(i, j), k, a, b]$ implies that $(x, k) \in$ Prev-Edge$_k$, and if $x \neq \text{start}$, then $R2(i, a, j, b)$ is compatible with at least one of $k$'s and one of $x$'s labels. If Prev-Support$[(i, j), k, a, b]$ becomes empty, then $R2(i, a, j, b)$ is no longer true in segments containing $i$, $j$, and $k$. |
| Next-Support$[(i, j), k, a, b]$ | $(i, x) \in$ Next-Support$[(i, j), k, a, b]$ means that $(k, x) \in$ Next-Edge$_k$, and if $x \neq \text{end}$, then $R2(i, a, j, b)$ is compatible with at least one of $k$'s and one of $x$'s labels. If Next-Support$[(i, j), k, a, b]$ becomes empty, then $R2(i, a, j, b)$ is no longer true in segments containing $i$, $j$, and $k$. |

Figure 29: Data structures and notation for MUSE PC-1.

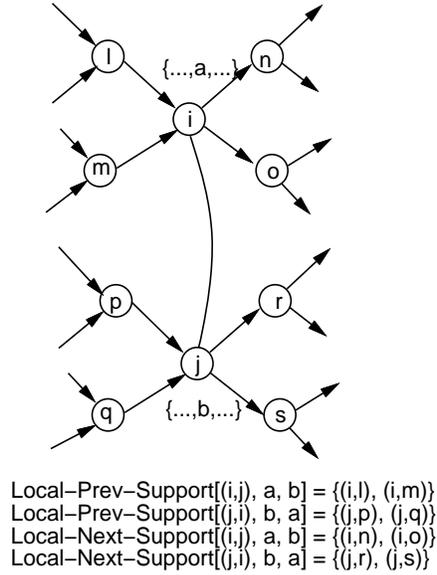

Local-Prev-Support[(i,j), a, b] = {(i,l), (i,m)}
Local-Prev-Support[(j,i), b, a] = {(j,p), (j,q)}
Local-Next-Support[(i,j), a, b] = {(i,n), (i,o)}
Local-Next-Support[(j,i), b, a] = {(j,r), (j,s)}

Figure 30: Local-Prev-Support and Local-Next-Support for the path consistency of an example DAG. The solid directed lines are members of $G$, and the solid undirected line represents the $(i, j)$ and $(j, i)$ members of $E$.

Support$[(j, i), k, b, a]$, Next-Support$[(i, j), k, a, b]$, and Next-Support$[(j, i), k, b, a]$ sets. Next-Support$[(i, j), k, a, b]$ includes all $(i, x)$ arcs which support $R2(i, a, j, b)$ given that there is a directed edge from $k$ to $x$, $R2(i, a, j, b) = 1$, $R2(i, a, k, c) = 1$, and $R2(k, c, j, b) = 1$ (Next-Support$[(j, i), k, b, a]$ is defined similarly). Prev-Support$[(i, j), k, a, b]$ includes all $(i, x)$ arcs which support $R2(i, a, j, b)$ given that there is a directed edge from $x$ to $k$, $R2(i, a, j, b) = 1$, $R2(i, a, k, c) = 1$, and $R2(k, c, j, b) = 1$ (Prev-Support$[(j, i), k, b, a]$ is defined similarly). Note that Prev-Support$[(i, j), k, a, b]$ will contain an ordered pair $(i, k)$ if $(i, k) \in$ Prev-Edge$_k$, and $(i, j)$ if $(j, k) \in$ Prev-Edge$_k$. Next-Support$[(i, j), k, a, b]$ will contain an ordered pair $(i, k)$ if $(k, i) \in$ Next-Edge$_k$ and $(i, j)$ if $(k, j) \in$ Next-Edge$_k$. These elements are included because the edge between those nodes is sufficient to allow the support. Dummy ordered pairs are also created to handle cases where a node is at the beginning or end of a network: when $(\text{start}, k) \in$ Prev-Edge$_k$, $(i, \text{start})$ is added to Prev-Support$[(i, j), k, a, b]$, and when $(k, \text{end}) \in$ Next-Edge$_k$, $(i, \text{end})$ is added to Next-Support$[(i, j), k, a, b]$.

## 4.2 The Running Time, Space Complexity, and Correctness of MUSE PC-1

The worst-case running time of the routine to initialize the MUSE PC-1 data structures (in Figure 32) is $O(n^3 l^3 + n^4 l^2)$, where $n$ is the number of nodes in a MUSE CSP and $l$ is the number of labels. Given that the number of $(i, j)$ elements in $E$ is $O(n^2)$ and the domain size is $O(l)$, there are $O(n^3 l^2)$ entries in the Counter array for which to determine the number of supporters, requiring $O(l)$ work; hence, initializing the Counter array requires $O(n^3 l^3)$ time. Additionally, there are $O(n^3 l^2)$ S sets to determine, each with $O(l)$ values, so the time required to initialize them is $O(n^3 l^3)$. Determining each Prev-Support$[(i, j), k, a, b]$

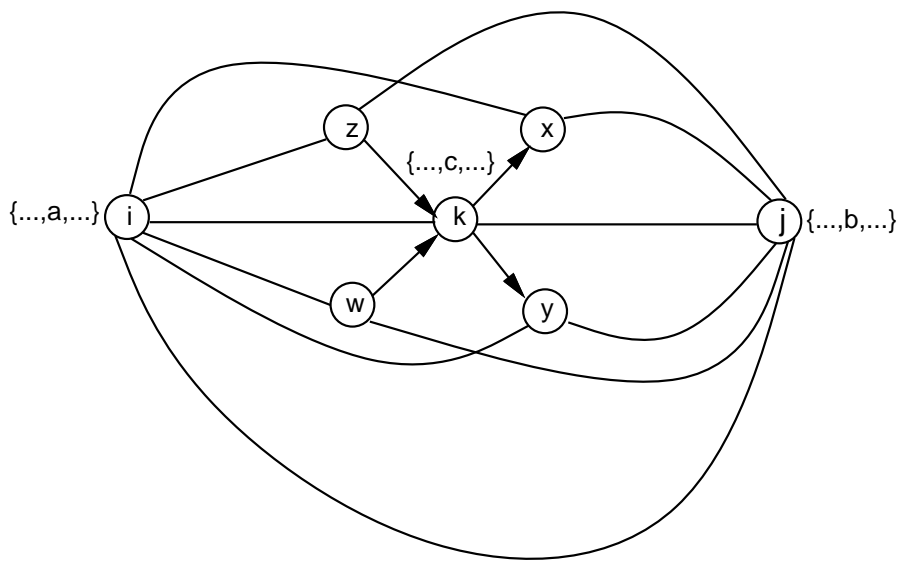

Figure 31: If it is found that Next-Edge$_k$ = $\{(k,x),(k,y)\}$, Counter$[(i,j),x,a,b]$ = 0, and Counter$[(i,j),y,a,b]=0$, then $R2(i,a,j,b)$ is ruled out for every segment containing $i$, $j$, and $k$. The solid directed lines are members of $G$, and the solid undirected lines represent members of $E$.

```
1.    List := φ;
2.    E := {(i, j)|∃σ ∈ Σ : i, j ∈ σ ∧ i ≠ j ∧ i, j ∈ N};
3.    for (i, j) ∈ E do
4.        for a ∈ L_i do
5.            for b ∈ L_j do {
6.                Local-Prev-Support[(i, j), a, b] := φ; Local-Next-Support[(i, j), a, b] := φ;
7.                for k ∈ N such that (i, k) ∈ E ∧ (j, k) ∈ E do {
8.                    S[(i, j), k, a, b] := φ;
9.                    M[(i, j), k, a, b] := 0;
10.                   Prev-Support[(i, j), k, a, b] := φ; Next-Support[(i, j), k, a, b] := φ;  }  }
11.   for (i, j) ∈ E do
12.       for a ∈ L_i do
13.           for b ∈ L_j such that R2(i, a, j, b) do {
14.               for k ∈ N such that (i, k) ∈ E ∧ (j, k) ∈ E do {
15.                   Total := 0;
16.                   for c ∈ L_k do
17.                       if R2(i, a, k, c) and R2(k, c, j, b) then {
18.                           Total := Total+1;
19.                           S[(i, k), j, a, c] := S[(i, k), j, a, c] ∪ {(j, b)}; }
20.                   if Total = 0 then {
21.                       List := List ∪ {[(i, j), k, a, b]};
22.                       M[(i, j), k, a, b] := 1; }
23.                   Counter[(i, j), k, a, b] := Total;
24.                   Prev-Support[(i, j), k, a, b] :=
                                    {(i, x)|(i, x) ∈ E ∧ (x = j ∨ (j, x) ∈ E) ∧ (x, k) ∈ Prev-Edge_k}
                                    ∪ {(i, k)|(i, k) ∈ Prev-Edge_k}
                                    ∪ {(i, start)|(start, k) ∈ Prev-Edge_k};
25.                   Next-Support[(i, j), k, a, b] :=
                                    {(i, x)|(i, x) ∈ E ∧ (x = j ∨ (j, x) ∈ E) ∧ (k, x) ∈ Next-Edge_k}
                                    ∪ {(i, k)|(k, i) ∈ Next-Edge_k}
                                    ∪ {(i, end)|(k, end) ∈ Next-Edge_k}; }
26.               Local-Prev-Support[(i, j), a, b] :=
                                    {(i, x)|(i, x) ∈ E ∧ (x = j ∨ (j, x) ∈ E) ∧ (x, i) ∈ Prev-Edge_i}
                                    ∪ {(i, start)|(start, i) ∈ Prev-Edge_i};
27.               Local-Next-Support[(i, j), a, b] :=
                                    {(i, x)|(i, x) ∈ E ∧ (x = j ∨ (j, x) ∈ E) ∧ (i, x) ∈ Next-Edge_i}
                                    ∪ {(i, end)|(i, end) ∈ Next-Edge_i}; }
```

Figure 32: Initialization of the data structures for MUSE PC-1.

1. **while** $List \neq \phi$ **do**
2.     Pop $[(i,j),k,a,b]$ from $List$;
3.     **for** $(k,c) \in \mathrm{S}[(i,j),k,a,b]$ **do** {
4.         Counter$[(i,k),j,a,c] := $ Counter$[(i,k),j,a,c] - 1$;
5.         Counter$[(k,i),j,c,a] := $ Counter$[(k,i),j,c,a] - 1$;
6.         **if** Counter$[(i,k),j,a,c] = 0 \wedge \mathrm{M}[(i,k),j,a,c] = 0$ **then** {
7.           $List := List \cup \{[(i,k),j,a,c],[(k,i),j,c,a]\}$;
8.           $\mathrm{M}[(i,k),j,a,c] := 1$; $\mathrm{M}[(k,i),j,c,a] := 1$; } }
9.         Update-Support-Sets$([(i,j),k,a,b])$; (see Figure 34) }

Figure 33: Eliminating inconsistent binary constraints in MUSE PC-1.

Update-Support-Sets $([(i,j),k,a,b])$ {
1.   **for** $(i,x) \in$ Prev-Support$[(i,j),k,a,b] \wedge x \neq j \wedge x \neq k \wedge x \neq$ start **do** {
2.     Prev-Support$[(i,j),k,a,b] := $ Prev-Support$[(i,j),k,a,b] - \{(i,x)\}$;
3.     Next-Support$[(i,j),x,a,b] := $ Next-Support$[(i,j),x,a,b] - \{(i,k)\}$;
4.     **if** Next-Support$[(i,j),x,a,b] = \phi \wedge \mathrm{M}[(i,j),x,a,b] = 0$ **then** {
5.       $List := List \cup \{[(i,j),x,a,b],[(j,i),x,b,a]\}$;
6.       $\mathrm{M}[(i,j),x,a,b] := 1$; $\mathrm{M}[(j,i),x,b,a] := 1$; } }
7.   **for** $(i,x) \in$ Next-Support$[(i,j),k,a,b] \wedge x \neq j \wedge x \neq k \wedge x \neq$ end **do** {
8.     Next-Support$[(i,j),k,a,b] := $ Next-Support$[(i,j),k,a,b] - \{(i,x)\}$;
9.     Prev-Support$[(i,j),x,a,b] := $ Prev-Support$[(i,j),x,a,b] - \{(i,k)\}$;
10.     **if** Prev-Support$[(i,j),x,a,b] = \phi \wedge \mathrm{M}[(i,j),x,a,b] = 0$ **then** {
11.       $List := List \cup \{[(i,j),x,a,b],[(j,i),x,a,b]\}$;
12.       $\mathrm{M}[(i,j),x,a,b] := 1$; $\mathrm{M}[(j,i),x,b,a] := 1$; } }
13.   **if** $(k,i) \in$ Prev-Edge$_i$ **then**
14.     Local-Prev-Support$[(i,j),a,b] := $ Local-Prev-Support$[(i,j),a,b] - \{(i,k)\}$;
15.   **if** Local-Prev-Support$[(i,j),a,b] = \phi$ **then** {
16.     $R2(i,a,j,b) := 0$; $R2(j,b,i,a) := 0$;
17.     **for** $(i,x) \in$ Local-Next-Support$[(i,j),a,b] \wedge x \neq j \wedge x \neq k \wedge x \neq$ end **do** {
18.       Local-Next-Support$[(i,j),a,b] := $ Local-Next-Support$[(i,j),a,b] - \{(i,x)\}$;
19.       **if** $\mathrm{M}[(i,j),x,a,b] = 0$ **then** {
20.         $List := List \cup \{[(i,j),x,a,b],[(j,i),x,b,a]\}$;
21.         $\mathrm{M}[(i,j),x,a,b] := 1$; $\mathrm{M}[(j,i),x,b,a] := 1$; } } }
22.   **if** $(i,k) \in$ Next-Edge$_i$ **then**
23.     Local-Next-Support$[(i,j),a,b] := $ Local-Next-Support$[(i,j),a,b] - \{(i,k)\}$;
24.   **if** Local-Next-Support$[(i,j),a,b] = \phi$ **then** {
25.     $R2(i,a,j,b) := 0$; $R2(j,b,i,a) := 0$;
26.     **for** $(i,x) \in$ Local-Prev-Support$[(i,j),a,b] \wedge x \neq j \wedge x \neq k \wedge x \neq$ start **do** {
27.       Local-Prev-Support$[(i,j),a,b] := $ Local-Prev-Support$[(i,j),a,b] - \{(i,x)\}$;
28.       **if** $\mathrm{M}[(i,j),x,a,b] = 0$ **then** {
29.         $List := List \cup \{[(i,j),x,a,b],[(j,i),x,b,a]\}$;
30.         $\mathrm{M}[(i,j),x,a,b] := 1$; $\mathrm{M}[(j,i),x,b,a] := 1$; } } } }

Figure 34: The function Update-Support-Sets$([(i,j),k,a,b])$ for MUSE PC-1.

| Approach | Nodes per Path | Degree of Node splitting | Number of Constraint Networks | Number of Nodes | Asymptotic Time |
|---|---|---|---|---|---|
| CSPs | $n$ | $k$ | $k^n$ | $n$ | $k^n n^3 l^3$ |
| MUSE CSP | $n$ | $k$ | 1 | $kn$ | $(kn)^3 l^3 + (kn)^4 l^2$ |

Table 2: Comparison of the space and time complexity for MUSE path consistency on a MUSE CSP to path consistency on multiple CSPs representing a node splitting problem (e.g., lexical ambiguity in parsing).

and Next-Support$[(i,j),k,a,b]$ requires O($n$) time, so the time required to calculate all Prev-Support and Next-Support sets is O($n^4 l^2$). Finally, the time needed to calculate all Local-Next-Support and Local-Prev-Support sets is O($n^3 l^2$) because there are O($n^2 l^2$) sets with up to O($n$) elements per set.

The worst-case running time for the algorithm which enforces MUSE path consistency (in Figures 33 and 34) also operates in O($n^3 l^3 + n^4 l^2$) time. Clearly there are O($n^3 l^2$) entries in the Counter array to keep track of in the algorithm. Each Counter$[(i,j),k,a,b]$ can be at most $l$ in magnitude, and it can never become negative, so the maximum running time for lines 4 and 5 in Figure 33 (given that elements, because of M, appear on the list only once) is O($n^3 l^3$). Because there are O($n^3 l^2$) Prev-Support and Next-Support lists, each up to O($n$) in size, the maximum running time required to eliminate O($n$) elements from those support sets is O($n^4 l^2$). Finally, since there are O($n^2 l^2$) Local-Next-Support and Local-Prev-Support sets from which to eliminate O($n$) elements, the worst-case time to eliminate items from the local sets is O($n^3 l^2$). Hence, the worst-case running time of the MUSE CSP path consistency algorithm is O($n^3 l^3 + n^4 l^2$).

The space complexity of MUSE CSP PC-1 is also O($n^3 l^3 + n^4 l^2$) because the arrays Counter and M contain O($n^3 l^2$) elements and there are O($n^3 l^2$) S sets, each containing O($l$) items; O($n^3 l^2$) Prev-Support and Next-Support sets, each containing O($n$) items; and O($n^2 l^2$) Local-Next-Support and Local-Prev-Support sets, each containing O($n$) items.

By comparison, the worst-case running time and space complexity for CSP path consistency, PC-4, is O($n^3 l^3$). Note that for applications where $\Sigma$ is representable as planar DAG or $l = n$, the worst-case running times of the algorithms are the same order. If we compare MUSE CSP to the use of multiple CSPs for problems where are $k$ alternative variables for a particular variable in a CSP, then MUSE CSP path consistency can be asymptotically more attractive, as shown in Table 2.

Because the proof of correctness for MUSE PC-1 is similar to our proof for MUSE AC-1, we will only briefly outline the proof here. A binary constraint looses support in MUSE PC-1 (see lines 16 and 25 in Figure 34) only if its Local-Prev-Support set or its Local-Next-Support set becomes empty (see lines 15 and 24 in Figure 34, respectively). In either case, it is inadmissible in any MUSE path consistent instance. We prove that a constraint's local support sets become empty if and only if it cannot participate in a MUSE path consistent instance of MUSE CSP. This is proven for Local-Next-Support (Local-Prev-Support follows by symmetry). Observe that if $R2(i,a,j,b) = 1$, and all of the nodes which immediately

$$\sigma_2 = \{ \, ① \, , ② \, , ③ \, \}$$
$$\sigma_1 = \{ \, ④ \, , ⑤ \, \}$$

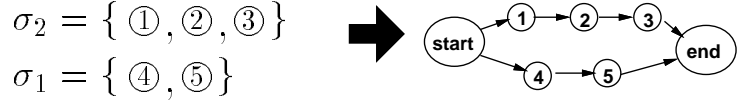

Figure 35: An example set of CSP problems which would not be a good candidate for MUSE CSP because of the lack of node sharing.

follow $i$ (and similarly $j$) in the DAG are incompatible with the truth of the constraint, then it cannot participate in a MUSE path consistent instance. In line 23 of Figure 34, $(i, k)$ is removed from the Local-Next-Support$[(i, j), a, b]$ only after $[(i, j), k, a, b]$ has been popped off $List$. The removal of $(i, k)$ from Local-Next-Support$[(i, j), a, b]$ indicates that the segment containing $i$, $j$, and $k$ does not support $R2(i, a, j, b)$. It remains to be shown that $[(i, j), k, a, b]$ is put on $List$ only if $R2(i, a, j, b)$ must be false for every segment which contains $i$, $j$, and $k$. This can be proven by induction on the number of iterations of the while loop in Figure 33 (much like the proof for MUSE AC-1). We must also show that if $R2(i, a, j, b) = 1$ after MUSE PC-1, then it is MUSE path consistent. For $R2(i, a, j, b)$ to be MUSE path consistent, there must exist at least one path from start to end which goes through nodes $i$ and $j$ such that all nodes $n$ on that path contain at least one label consistent with the constraint. This proof would be similar to the second half of the proof for MUSE AC-1 correctness. From this, we may conclude that MUSE PC-1 builds the largest MUSE path consistent structure.

## 5. Combining CSPs into a MUSE CSP

Problems which have an inherent lattice structure or problems which can be solved by the node splitting approach are natural areas of application for MUSE CSP, because an exponential number of CSPs are replaced by a single instance of MUSE CSP, and the DAG representation of $\Sigma$ is inherent in the problem. In this section we discuss DAG construction for other application areas which would benefit from the MUSE CSP approach, but for which it is not as obvious how to construct the DAG. Any set of CSP problems can be used as the segments of a MUSE CSP. For example, Figure 35 illustrates how two instances of a CSP can be combined into a single MUSE CSP. However, using MUSE CSP for this example would not be the right choice; node sharing cannot offset the cost of using the extra MUSE AC-1 data structures.

Multiple nodes which have the same name in various CSPs can potentially be represented as a single node in a MUSE CSP. We assume that if two nodes, $k_1$ and $k_2$ are given the same name (say $k$) in two instances of CSP, then they have the same domain and obey the same constraints, i.e.:

1. $L_{k_1} = L_{k_2}$ (i.e., their domains are equal.)

2. $R1(k_1, a) = R1(k_2, a)$ for every $a \in L_{k_1}, L_{k_2}$ (i.e., their unary constraints are the same.)

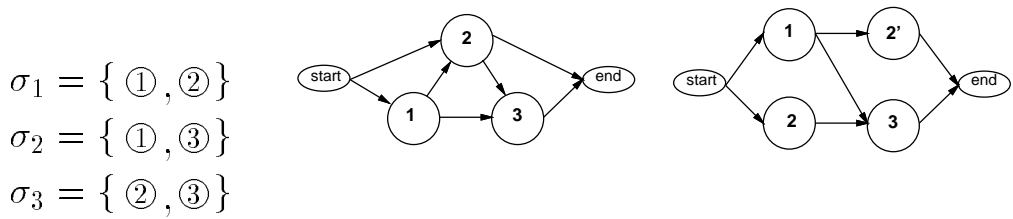

$$\sigma_1 = \{\;①\,,②\;\}$$
$$\sigma_2 = \{\;①\,,③\;\}$$
$$\sigma_3 = \{\;②\,,③\;\}$$

Figure 36: An example of how maximal node sharing leads to spurious segments. The first DAG contains two paths, {1,2,3} and {2}, which correspond with none of the segments. The second DAG presents a preferred sharing as created by the CREATE-DAG routine.

3. $R2(k_1, a, i, b) = R2(k_2, a, i, b)$ for labels $a \in L_{k_1}, L_{k_2}$ and $b \in L_i$, where $i$ is in both segments (i.e., their binary constraints are the same.)

However, as illustrated in Figure 36, too much sharing of common nodes can introduce additional segments that do not appear in the original list of CSPs. Because the extra segments can cause extra work to be done, it is often desirable to create a DAG which shares nodes without introducing extra segments. The algorithm CREATE-DAG, shown in Figure 38 takes an arbitrary set of CSP problems as input (a list of segments), and outputs a DAG representation for those CSPs which shares nodes without introducing spurious segments. CREATE-DAG calls an auxiliary procedure ORDER-SIGMA defined in Figure 39. The data structures used in these two routines are defined in Figure 37.

To hold the individual segments in $\Sigma$, the routine CREATE-DAG uses a special data structure for ordered sets which supports some useful operations. If $\sigma$ is a segment and $n$ is an integer, then $\sigma[n]$ is the node at position $n$ in $\sigma$. $\sigma[0]$ is always the start node, and $\sigma[|\sigma| - 1]$ is always the end node. $\sigma[k..m]$ is the ordered subset of $\sigma$ consisting of all the nodes in positions $k$ through $m$. In addition, the ordered set allows us to insert a node $i$ immediately after any node $j$ which is already in the set. Each node $\sigma[pos]$ is a structure with a name field and a next-set field, which is the set of names of nodes that follow the node $\sigma[pos]$ in a set of segments.

CREATE-DAG begins by adding special purpose start and end nodes to each segment. It then calls the routine ORDER-SIGMA shown in Figure 39 to order the nodes in each segment. ORDER-SIGMA orders the nodes of each segment such that the ones that are the most common tend to occur earlier in the set. To order the elements, it uses the operator $>$ (i.e., larger than) which is defined on nodes. Note that the start node is defined to be the "largest" node, and the end node the "smallest" node. In addition, $i > j$ means either that $i$ appears in more segments than $j$ does, or if they both appear in the same number of segments, then $i$ has a lower ordinal number than $j$. Thus the operator $>$ induces a total ordering on the nodes in $N$.

When ORDER-SIGMA is first called by CREATE-DAG it selects the largest node $i$ which is smaller than the start node. It then constructs the set $S$, which is a set of those segments containing $i$. At this point, the segments in $S$ are ordered such that the start node is first and $i$ is second. It then calls ORDER-SIGMA to order $S$ for nodes smaller than $i$. Once the recursive call is done, any segments that were not in $S$ are considered (i.e., $Z$). Note that

| Notation | Meaning |
|----------|---------|
| $\Sigma$ | A set of node sets. Each node set represents a CSP. |
| $\sigma$ | $\sigma$ is a node set or segment in $\Sigma$. Each $\sigma$ set is modified to include begin and end nodes for the Create-DAG algorithm to work properly. Note that $\sigma[0]$ is always the start node, and $\sigma[|\sigma|-1]$ is always the end node. Each node $\sigma[pos]$ is a structure with a name and a next-set (names of nodes that follow the node in the DAG). |
| $G$ | $G$ is the set of node pairs $(i, j)$ such that there exists a directed edge from $i$ to $j$ in the DAG created by Create-DAG. |
| $N$ | $N$ is the set of nodes that have been placed in the DAG by Create-DAG. |
| $Z$ | $Z$ is a set of segments to order with respect to node $j$ in Order-Sigma. |
| $j$ | The node $j$ is used in Order-Sigma to order the remaining elements which are smaller than that node. |
| $U$ | $U$ is a set of nodes already considered in the current call to Order-Sigma. |
| $R$ | $R$ is a set of nodes in $Z$ in Order-Sigma. |
| $i$ | The node $i$ is the largest node which is smaller than $j$ in $R - U$ (if non-empty) or $R$ in Order-Sigma. |
| $S$ | In Order-Sigma, $S$ is a set of segments in $Z$ which contain node $i$. |

Figure 37: Data structures used in Create-DAG and Order-Sigma.

CREATE-DAG () {
1.    Add start as the first node and end as the last node for every segment in $\Sigma$;
2.    ORDER-SIGMA($\Sigma$, start);
3.    **for** pos := 1 **to** maximum segment length {
4.        $\Sigma' := \text{copy}(\Sigma)$;
5.        **for** $\sigma \in \Sigma' \wedge |\sigma| - 1 > \text{pos}$ {
6.            **if** $\sigma[\text{pos}].\text{name} = \text{end}$ **then** {
7.                $G := G \cup \{(\sigma[\text{pos} - 1], \sigma[\text{pos}])\}$; }
8.            **else** {
9.                SAME_EDGE_SET := $\{\sigma 1|\ \sigma 1[\text{pos} - 1].\text{name} = \sigma[\text{pos} - 1].\text{name} \wedge$
                                    $\sigma 1[\text{pos}].\text{name} = \sigma[\text{pos}].\text{name}\}$;
10.               next-set := $\{\sigma[\text{pos} + 1].\text{name} \,|\sigma \in \text{SAME\_EDGE\_SET}\}$;
11.               $\Sigma' := \Sigma' - \text{SAME\_EDGE\_SET}$;
12.               **if** a node with $\sigma[\text{pos}].\text{name}$ is not in $N$ **then** {
13.                   $N := N \cup \sigma[\text{pos}]$;
14.                   $\sigma[\text{pos}].\text{next} := \text{next-set}$;
15.                   $G := G \cup \{(\sigma[\text{pos} - 1], \sigma[\text{pos}])\}$; }
16.               **else** {
17.                   node := get the node in $N$ with the name $\sigma[\text{pos}].\text{name}$;
18.                   **if** node.next = next-set **then** {
19.                       $G := G \cup \{(\sigma[\text{pos} - 1], \sigma[\text{pos}])\}$; }
20.                   **else** {
21.                       new-node : = Create a new node;
22.                       new-node.name := concatenate($\sigma[\text{pos}].\text{name}$, ');
23.                       node := get the node in $N$ named new-node.name (if there is one);
24.                       **while** node && node.next !=next-set **do** {
25.                           new-node.name := concatenate(new-node.name, ');
26.                           node := get the node in $N$ named new-node.name (if there is one); }
27.                       **if** (node = NULL) **then**
28.                           $N := N \cup \text{new-node}$;
29.                       **else** new-node := node;
30.                       new-node.next := next-set;
31.                       Replace $\sigma[\text{pos}].\text{name}$ with new-node.name in $\sigma[\text{pos} - 1].\text{next}$;
32.                       $G := G \cup \{(\sigma[\text{pos} - 1], \text{new-node})\}$;
33.                       Replace every occurance of $\sigma[\text{pos}]$ at pos with new-node
                            in all segments of SAME_EDGE_SET; } } } } }
34.   Eliminate start and end from $G$ and from each $\sigma \in \Sigma$; }

Figure 38: Routine to create a DAG to represent $\Sigma$.

```
Order-Sigma (Z, j) {
1.      U := φ;
2.      while Z ≠ φ {
3.          R := ⋃ σ;
               σ∈Z
4.          if R − U ≠ φ then
5.              i := the "largest" node in R − U which is less than j ;
6.          else
7.              i := the "largest" node in R which is less than j ;
8.          S := {σ|σ ∈ Z ∧ i ∈ σ} ;
9.          Z := Z − S;
10.         if i ≠ end then {
11.             for σ ∈ S {
12.                 Put i after j in σ;
13.                 U := U ∪ σ; }
14.             Order-Sigma(S, i) } } }
```

Figure 39: The routine to arrange the nodes within the segments for convenient merging.

after the first iteration of the loop, there is a preference to select the largest node that was not contained in the segments that were ordered by the recursive call to Order-Sigma. These items are independent of the ordered segments, and so will not create spurious paths when placed early in the DAG; however, items that occur in the already ordered segments, if placed earlier than items that do not occur in the ordered segments would tend to introduce spurious paths. The while loop continues until all segments in $\Sigma$ are ordered. The worst-case running time of Order-Sigma is $O(n^2)$, where $n$ is the sum of the cardinalities of the segments in $\Sigma$.

Once Order-Sigma orders the nodes in the segments, Create-DAG begins to construct the DAG, which is represented as a set of nodes $N$ and a set of directed edges $G$. The DAG is constructed by going through each segment beginning with the position of the second element (the position after start). The for loop on line 3 looks at nodes in a left to right order, one position at a time, until all the elements of each segment have been added to $G$. If a node with a certain name has not already been placed in $N$ (i.e., the set of nodes already in the DAG being created) then adding the node to the graph (as well as a directed edge between $\sigma[pos-1]$ and $\sigma[pos]$ to $G$) cannot create any spurious paths in the DAG. On the other hand, if a node with the same name as $\sigma[pos]$ had already been placed in $N$, then it is possible that the current segment could add paths to the DAG that do not correspond to any of the segments in $\Sigma$. To avoid adding spurious segments, we deal with all segments at one time that share the same previous node and have a node with the same name at the current position. The basic idea is to add that edge only once and to keep track of all nodes that can follow that node in the DAG. By doing this, we can easily determine whether that same node can be used if it occurs in another segment in a later position. The same node can be used only if it is followed by precisely the same set of next nodes that follow the node already placed in the graph; otherwise, the second node would have to be renamed to avoid adding spurious segments. In such an event, we create a new name for the node.

285

Note that once the DAG is complete, we eliminate the start and end nodes from $G$ (and their corresponding outgoing and incoming edges) to make $G$ consistent with its use in the MUSE arc consistency and MUSE path consistency algorithms. The running time of CREATE-DAG is also $O(n^2)$, where $n$ is the sum of the cardinalities of the segments in $\Sigma$.

Even though the DAGs produced by the routine CREATE-DAG do have nice properties, this routine should probably be used only as a starting point for custom combining routines which are specific to the intended application area. We believe that domain-specific information can play an important role in MUSE combination. An example of a domain specific combining algorithm is presented in (Harper et al., 1992), which describes a spoken-language parsing system which uses MUSE CSP. A distinguishing feature of this application's combining algorithm is that instead of avoiding the creation of extra segments, it allows controlled introduction of extra segments because the extra segments often represent sentences which an N-Best sentence spoken language recognition system would miss.

## 6. Conclusion

In conclusion, MUSE CSP can be used to efficiently represent several similar instances of the constraint satisfaction problem simultaneously. If multiple instances of a CSP have some common variables with the same domains and compatible constraints, then they can be combined into a single instance of a MUSE CSP, and much of the work required to enforce node, arc, and path consistency need not be duplicated across the instances, especially if the constraints are sufficiently tight.

We have developed a MUSE CSP constraint-based parser, PARSEC (Harper & Helzerman, 1995a; Harper et al., 1992; Zoltowski et al., 1992), which is capable of parsing word graphs containing multiple sentence hypotheses. We have developed syntactic and semantic constraints for parsing sentences, which when applied to a word graph, eliminate those hypotheses that are syntactically or semantically incorrect. For our work in speech processing, the MUSE arc consistency algorithm is very effective at pruning the incompatible labels for the individual CSPs represented in the composite structure. When extracting each of the parses for sentences remaining in the MUSE CSP after MUSE AC-1, it is usually unnecessary to enforce arc consistency on the CSP represented by that directed path through the network because of the tightness of the syntactic and semantic constraints.

Speech processing is not the only area where segmenting the signal into higher-level chunks is problematic. Vision systems and handwriting analysis systems have comparable problems. In addition, problems that allow for parallel alternative choices for the type of a variable, as in parsing lexically ambiguous sentences, are also excellent candidates for MUSE CSP.

C++ implementations of the algorithms described in this paper are available at the following location: ftp://transform.ecn.purdue.edu/pub/speech/harper_code/. This directory contains a README file and a file called muse_csp.tar.Z.

## Acknowledgements

This work was supported in part by the Purdue Research Foundation and a grant from Intel Research Council. We would like to thank the anonymous reviewers for their insightful recommendations for improving this paper.